\documentclass[11pt]{article}

\usepackage{acl}
\usepackage{hyperref}
\usepackage{times}
\usepackage{latexsym}
\usepackage{amsmath}
\usepackage{amsfonts}
\usepackage{subfigure}
\usepackage{xspace}
\usepackage{paralist}
\usepackage{listings}
\usepackage[T1]{fontenc}

\usepackage[utf8]{inputenc}

\usepackage{microtype}

\usepackage{inconsolata}

\usepackage{graphicx}

\title{Beyond Memorization: Extending Reasoning Depth with Recurrence, Memory and Test-Time Compute Scaling}

\author{
\parbox{\textwidth}{\centering
\textbf{
Ivan Rodkin$^{1,2}$ \quad
Daniil Orel$^{1}$ \quad
Konstantin Smirnov$^{1}$ \quad
Arman Bolatov$^{1}$
}\\
\textbf{
Bilal Elbouardi$^{1}$ \quad
Besher Hassan$^{1}$ \quad
Yuri Kuratov$^{2,3}$ \quad
Aydar Bulatov$^{2,3}$
}\\
\textbf{
Preslav Nakov$^{1}$ \quad
Timothy Baldwin$^{1}$ \quad
Artem Shelmanov$^{1}$ \quad
Mikhail Burtsev$^{4}$
}\\[0.5em]
{\normalfont
$^{1}$MBZUAI \quad
$^{2}$MIRAI \quad
$^{3}$Cognitive AI Systems Lab \quad \\
$^{4}$London Institute for Mathematical Sciences
}\\[0.3em]
{\normalfont\ttfamily
\{ivan.rodkin,daniil.orel,konstantin.smirnov,arman.bolatov\}@mbzuai.ac.ae
}\\
{\normalfont\ttfamily
artem.shelmanov@mbzuai.ac.ae \quad mb@lims.ac.uk
}
}
}

\newcommand{\gptneox}{GPTNeox\xspace}
\begin{document}
\maketitle
\begin{abstract}
Reasoning is a core capability of large language models (LLMs), yet how multi-step reasoning is learned and executed remains unclear. We study this question in a controlled cellular-automata (1dCA) framework that excludes memorization by using disjoint training and test rules. Given a short state sequence, the model is required to \emph{infer} the hidden local rule and then \emph{chain} it to predict multiple future steps. Our evaluation shows that LLMs largely fail to reliably solve a natural-language proxy of the proposed task. We find that most neural architectures trained from scratch can learn rule inference and achieve high next-step accuracy, but performance drops sharply as the required number of intermediate reasoning steps increases. Experiments show that increasing model depth is crucial, and extending \emph{effective} depth via recurrence, memory, or test-time compute improves results but remains bounded. The code is available on github: \href{https://github.com/RodkinIvan/associative-recurrent-memory-transformer/tree/ACT}{https://github.com/RodkinIvan/associative-recurrent-memory-transformer/tree/ACT}.
\end{abstract}

\section{Introduction}
Large Language Models (LLMs) demonstrate impressive capabilities in problem-solving and reasoning tasks\footnote{\href{https://x.com/OpenAI/status/1954969035713687975}{https://x.com/OpenAI/status/1954969035713687975}} \citep{openai2024learning, guo2025deepseek, deepmind_gemini_deep_think_imo_2025}. However, extensive evidence from ongoing research shows that LLMs still face challenges in multi-step reasoning~\cite{dziri2024faith,wan2024b,holliday2024conditional,gandarela2024inductive,mondorf2024liar,shojaee2025illusionthinkingunderstandingstrengths} and planning~\cite{valmeekam2024llms}, particularly when required to infer and apply underlying rules from data. 
These observations raise the following questions:
\\
\\
\begin{enumerate}
    \item \textit{Is the reasoning exhibited by LLMs the result of genuine generalization or merely memorization?}
    \item \textit{How does task difficulty scale as the required number of reasoning steps increases?}
    \item \textit{To what extent do a model's architectural inductive biases, training objectives, and inference procedures limit its reasoning capabilities?}
\end{enumerate}

Transformers~\citep{vaswani2017attention} are universal function approximators, and with unbounded depth and precision, they are Turing-complete~\citep{cybenko1989approximations,hornik1989multilayer,dehghani2018universal-transformer,yun2019transformers,bhattamishra2020computational,perez2021attention,sanford2024representational}. Yet, \emph{finite-depth, fixed-width} models used in practice cannot process arbitrarily long inputs in a single forward pass, and they provably fail on tasks such as graph connectivity, Boolean formula evaluation, and exact arithmetic beyond a bounded length~\citep{merrill2022saturated,merrill2023parallelism,strobl2024formal,feng2024towards}.

One way to sidestep this depth barrier is to let the model \emph{write a scratch-pad} of intermediate tokens. Chain-of-Thought (CoT) prompting, process supervision, and reinforcement learning (RL) encourage models to emit multi-step rationales before producing the final answer~\citep{wei2022chain,uesato2022solving,wang2023math,yao2024tree,kumar2024training}. Generating and consuming these extra tokens increases the computational depth with the rationale length, enabling transformers to solve dynamic-programming benchmarks~\citep{feng2024towards} and to recognize regular languages with linear decoding depth~\citep{merrill2024expressive}. The main drawback is the need for supervision over intermediate steps, which is expensive or might be unavailable.

A complementary avenue is to \emph{recycle hidden states}. Segment-level recurrence in memory-augmented transformers~\citep{weston2014memory,graves2014neural} enables the re-feeding of hidden states across segments~\citep{dai2019transformerxl,raecompressive2019,rmt_2022,chevalier2023adapting,armt}, whereas state-space models achieve long-range interactions by leveraging linear dynamical systems~\citep{gu2021s4,gu2023mamba}. Recurrence deepens the network without emitting extra tokens, but the number of recurrent steps is still limited by the input length. \emph{Adaptive Computation Time} (ACT)~\citep{graves2016adaptive} removes this upper bound entirely: the model learns to allocate a variable number of layer updates to each token, halting once further computation is predicted to be unhelpful.  In principle, ACT grants transformers \emph{unbounded effective depth} while preserving parameter efficiency, which is an appealing property for reasoning tasks that require widely varying amounts of computation.

In this paper, we study \emph{rule abstraction} and \emph{multi-step reasoning} in neural models using a controlled 1D Cellular Automata (1dCA) setting that prevents memorization by holding out rule sets between training and testing. We cast reasoning as variable-horizon prediction and quantify how architectures and depth-extension strategies cope as the look-ahead \(k\) increases. Our contributions are:

\begin{figure*}[t]
  \centering
  \includegraphics[width=1\textwidth]{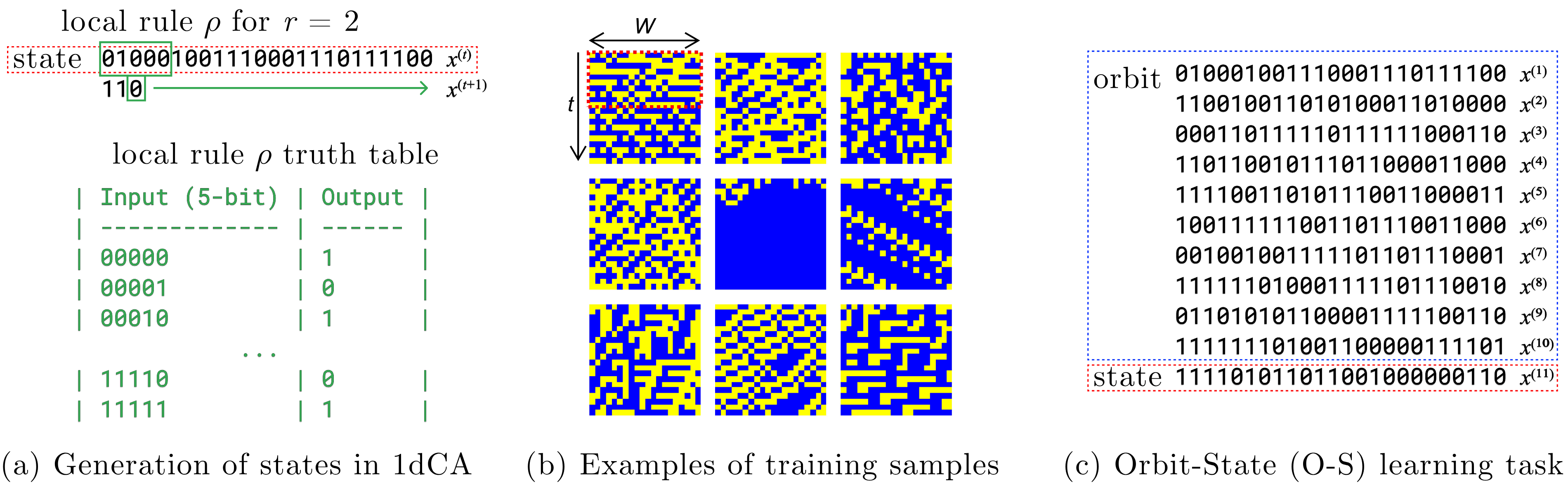}
  \caption{\textbf{Learning One-dimensional Cellular Automata (1dCA).}
  \textbf{(a)} Update of state with local rule.  
  \textbf{(b)}~ Orbit of 1dCA is a sequence of binary strings of size $W = 20$. The first $k = 10$ states marked by the red rectangle encode transformer input.
  \textbf{(c)} Given a part of the orbit a model learns to predict the next state (O-S).}
  \label{fig:methods}
\end{figure*}

\begin{itemize}
    \item \textbf{\textsc{1dca-Reasoning} benchmark.} A variable-complexity dataset that disentangles rule induction from state propagation; train/test rule sets are disjoint to preclude memorization.
    \item \textbf{LLMs evaluation in natural language.} A new \emph{Handsup} task -- a worded proxy equivalent to the 1dCA task -- used to assess LLMs under varying look-ahead and rule complexity, showing that many LLMs fail on the simplest radius-1 setting when they reason in natural language.
    \item \textbf{Comprehensive architectural comparison:} Side-by-side evaluation of small models including Transformers (GPT-NeoX), LSTMs, state-space models (Mamba), and a memory-augmented Transformer (AR\noindent MT) under identical conditions. Fixed-depth (4-layer) models solve \(k{=}1\) but collapse for \(k{\ge}2\); ARMT extends to \(k{=}2\). We corroborate these trends on a group-multiplication benchmark \citep{merrill2024illusion}.
    \item \textbf{Depth-extension analysis.} With 4-layer backbones: (i) Adaptive Computation Time (ACT) reliably adds \(\sim{+}1\) an effective step to the transformer-based architectures with modest compute; (ii) GRPO (RL) reaches \(k{=}3\) \emph{without} intermediate supervision; and (iii) token-level CoT attains near-perfect accuracy up to \(k{=}4\).
\end{itemize}

\section{Methods}
\label{sec:methods}


\paragraph{Modeling Reasoning with 1d Cellular Automata.}
Reason is the capacity to consciously apply logic by drawing valid conclusions from new or existing information.\footnote{\url{https://en.wikipedia.org/wiki/Reason}} Reasoning about an unfamiliar process naturally splits into two parts: (i) inferring the hidden law that drives state transitions and (ii) chaining that law to predict multiple future steps. One-dimensional cellular automata (1dCA) provide a minimal, fully observable sandbox for this: a local Boolean rule—the toy universe’s “micro-physics’’—updates each binary state from its neighborhood. In our benchmark, the rule is withheld, and the train/test rule sets are disjoint, so rote lookup cannot succeed. To solve a task, the model must first induce the rule from observed orbits and then apply it repeatedly to roll out future states, cleanly separating genuine rule-based reasoning from mere memorization.

\paragraph{Background.}
A \textit{One-dimensional Cellular Automaton (1dCA)} is a one-dimensional dynamical system in which space and time are discrete. Let $r \in \mathbb{N} : r \geq 1 $ be the \textit{neighborhood radius} in the space represented by a regular lattice of $W \in \mathbb{N} : W \geq 2r + 1 $ identical, locally-interconnected \textit{cells} with binary state spaces, $\mathbb{S} = \{0, 1\} $. The 1dCA's \textit{global state}, $x \in \mathbb{S}^W $, is a lattice configuration specified by the values of all states of all cells in the lattice at a given time. This state evolves deterministically in synchronous, discrete time steps according to a \textit{global map} $g_{\rho} : \mathbb{S}^W \rightarrow \mathbb{S}^W $ defined by a \textit{local rule} $\rho : \mathbb{S}^{2r+1} \rightarrow \mathbb{S} $, so $ [g_{\rho}(x)]_w = \rho(x_{w-r}, \ldots, x_w, \ldots, x_{w+r})$ (Fig.\ref{fig:methods}a). The sequence of states a 1dCA passes through during its \textit{space–time evolution}, $\mathcal{O}^T(x) = [x, g_{\rho}(x), g_{\rho}(g_{\rho}(x)), \ldots, g_{\rho}^{T-1}(x)] $, defines its \textit{trajectory} or \textit{orbit} from an \textit{initial condition} (configuration) $x $ for $T \in \mathbb{N} : T \geq 1 $. Examples of 1dCA orbits are visualized in \autoref{fig:methods}(b). The generated dataset can be found on Hugging Face: \href{https://huggingface.co/datasets/irodkin/1dCA_r2s20T20}{irodkin/1dCA\_r2s20T20}.

\paragraph{Benchmark for reasoning.}
Our benchmark instantiates multi-step reasoning with 1dCA trajectories: each example provides an orbit (e.g., 10 states) generated by a hidden rule; training and test use disjoint rule sets. The model must infer the rule from the observed states and predict configurations, forcing it to learn a rule-inference procedure rather than memorize instance-specific mappings. We vary difficulty via \emph{look-ahead} prediction: to give $g_{\rho}^{T+k-1}(x)$ for $k\in{1,2,3,4}$ steps ahead (without intermediate states), the model must internally roll out the dynamics, chaining the inferred rule. We call $k$ the \emph{depth of reasoning} and study which architectures can achieve depth under this setting.

\paragraph{Task variants.}
The benchmark could emulate situations where we have supervision on intermediate steps (i.e., LLM's thinking process) and when we only have a final look-ahead state. We consider two variations of tasks designed to assess different aspects of predictive modeling and rule inference:

\textit{Orbit-State (O-S)}: given an orbit $\mathcal{O}^T(x) = [x^{(1)}, x^{(2)}, \ldots, x^{(T)}]$ where $x^{(1)} \in \mathbb{S}^W$, the objective is to predict the state $x^{(T+k)}$ at look-ahead $k \in \mathbb{N} : k \geq 1 $. For $k=1$ (see Fig.\ref{fig:methods}c) this is a single-step prediction simulating an elementary act of reasoning. For $k>1$ multiple intermediate inference steps are required for the answer.

\textit{Orbit-Orbit (O-O)}: given an orbit $\mathcal{O}^T(x)$ for some $k>1$ predict the subsequent states up to time $T+k$, generating $\mathcal{O}_{T+1}^{T+k}(x) = [x^{(T+1)}, \ldots, x^{(T+k)}]$. This task simulates step-by-step multi-step reasoning as a learning objective. It's important to note that the predicted states are \textbf{not} fed into the model to predict the next ones; instead, all $k$ states are predicted from the corresponding number of mask tokens.

\begin{figure*}[t]
\centering
\includegraphics[width=\textwidth]{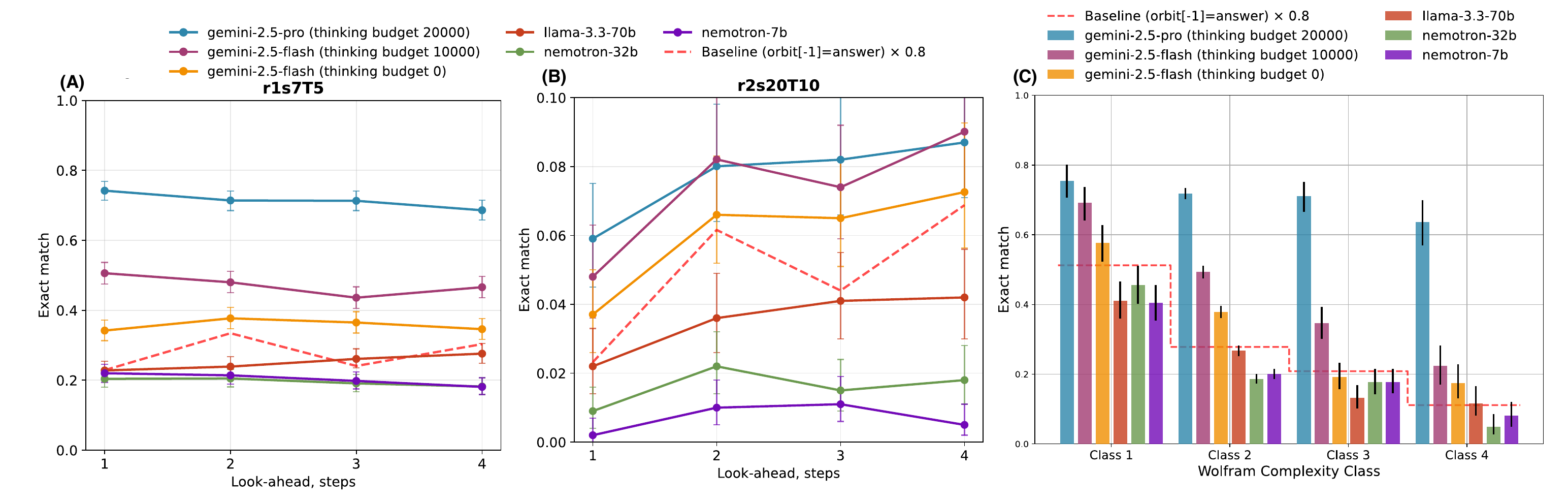}
\caption{\textbf{Large Language Models struggle to solve reasoning 1dCA style tasks in natural language game.} \textbf{(a)} Only Gemini-2.5-pro achieves notable performance in predicting the next state in Handsup game with $r{=}1$ and $w{=}7$ players given history of $T{=}5$ rounds. \textbf{(b)} None of models demonstrate reasonable scores (above 10\%) for harder game with more distant dependencies ($r{=}2$), more friends (20) and longer history (10). Only Gemini models achieve scores higher than the \textit{baseline} score, where the round is predicted the same as the last known round. \textbf{(c)} shows performance on the simple handsup game across different subsets of the dataset, split with respect to the Stephen Wolfram's rule classification of ECA. We observe the performance degradation on the tasks with rules of higher complexity (Class 3 and 4). All values are mean exact match with $95\%$ CIs. The final game state was extracted from the models' answers with Gemma3-12B-IT model with $\approx 80\%$ EM extraction accuracy  so we scale the baseline accordingly.}
\label{fig:llm-handsup}
\end{figure*}

\begin{figure*}[t]
    \centering
    \includegraphics[width=\textwidth]{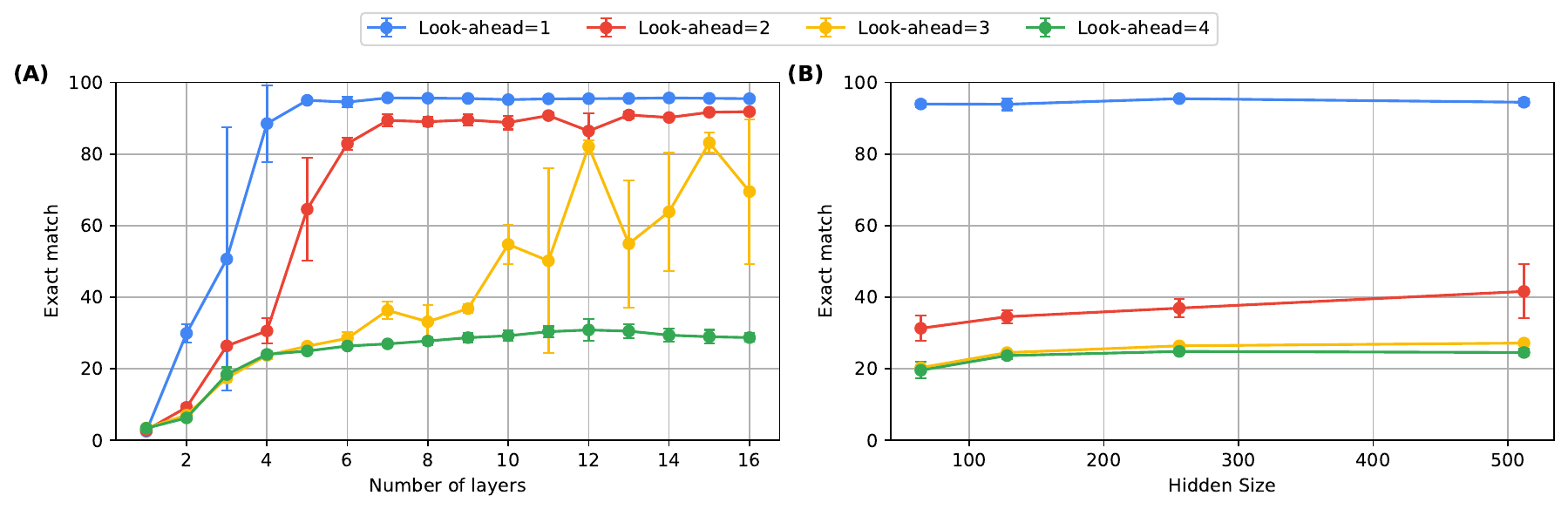}
    \caption{\textbf{Depth — not width — drives multi-step accuracy.}
    Exact-match accuracy for look-ahead horizons \(k\in\{1,2,3,4\}\) as a function of \textbf{(a)} transformer layer count and \textbf{(b)} embedding dimension \(d_{\text{model}}\).
    Deeper networks boost performance sharply for \(k{\ge}2\) and plateau beyond six layers, whereas widening the model yields only marginal gains across all horizons.}
    \label{fig:depth_width}
\end{figure*}

\paragraph{Neural Models.}
In our study, we consider LLMs and small models from several widely applied architectural families. Long Short-Term Memory (LSTM) networks \cite{hochreiter1997long}, a class of recurrent neural networks (RNNs), have proven effective at capturing sequential dependencies in NLP tasks. However, their inherent sequential processing limits efficiency and scalability. 

Transformers \cite{vaswani2017attention} address these limitations by processing entire input sequences simultaneously through self-attention, enabling parallel computation and better handling of long-range dependencies compared to RNN-based models. 

State space models (SSMs) \cite{gu2021s4} offer an alternative approach to sequence modeling by leveraging structured state representations and computationally efficient recurrence mechanisms across varied lengths, temporal scales, and regimes.
We consider the Associative Recurrent Memory Transformer (ARMT) \cite{armt}, an extension of the transformer designed to enhance memory capabilities. ARMT builds on the Recurrent Memory Transformer~\cite{rmt_2022} by incorporating quasi-linear attention mechanisms that improve information transfer across input blocks, mitigating limitations in long-context processing. We discuss the properties of these models in \autoref{appx:models_disc}.


We also explore several approaches to enhance reasoning in neural networks, including Chain-of-Thought, Reinforcement Learning (RL) methods, Group Relative Policy Optimization (GRPO), and Adaptive Computation Time.

\paragraph{Chain-of-Thought (CoT)} prompting \cite{wei2022chain} is a  technique for enhancing the reasoning capabilities of LLMs. 
Unlike  standard prompting techniques, which attempt to directly infer an answer from the input, CoT forces the model to explicitly generate intermediate reasoning steps while solving a problem, allowing it to reference these tokens as a form of recurrent state. This mechanism effectively increases the formal computational power of the model~\cite{merrill2024expressive} and extends its effective depth, enabling LLMs to perform multi-step reasoning, particularly in tasks such as mathematical problem-solving, logical inference, and commonsense reasoning~\cite{wei2022chain}. 
In this work, we refer to this technique when we explicitly train the model to generate step-by-step reasoning traces.

\paragraph{Learning to reason with RL.}
Another common practice involves training LLMs with reinforcement learning methods such as proximal policy optimization (PPO)~\cite{schulman2017proximal} and group relative policy optimization (GRPO)~\cite{shao2024deepseekmath} after supervised finetuning in order to improve the generation of reasoning traces. RL post-training has been shown to improve instruction following~\cite{ouyang2022training} as well as mathematical~\cite{wang2023math} and general reasoning performance in LLMs~\cite{havrilla2024teaching,kumar2024training, guo2025deepseek}. 

Compared to supervised methods, training to reason with GRPO requires no supervision on intermediate reasoning steps. It only relies on rewards from correct final answers and maintaining the desired format.

\paragraph{Adaptive Computation Time (ACT)}~\citep{graves2016adaptive} is the mechanism proposed to allow recurrent and self-attentive models to perform a variable number of computation steps within each time-step dynamically. The core idea is to enable different parts of the sequence to have different computational complexities, which is particularly useful for tasks with non-uniform requirements for computation. In this class of models, a halting unit dynamically decides how much ``thinking time'' should take place at each step, thus adaptively scaling the effective reasoning depth of the model. Prior work demonstrates that reusing model parameters for looped computation can yield a two- to three-fold increase in parameter efficiency \citep{zhu2025scalinglatentreasoninglooped}. For mathematical formulation, check the \autoref{appx:act}.

\paragraph{Recurrent Memory Transformers.} Recurrent Memory Transformers (RMT) offer a trade-off between expressive recurrent models and efficiently trainable transformers. It augments a transformer with recurrent steps between fixed-size segments, while tokens within each segment are still processed in parallel~\citet{rmt_2022}.

Recurrence is implemented by passing the output of special memory tokens from one segment to the input of the next segment. An enhanced variant, the Associative Recurrent Memory Transformer (ARMT) \citep{armt}, performs these recurrent updates using quasi-linear attention in each transformer layer. In this work, we use ARMT as a representative of recurrent memory transformers.

\begin{figure*}[t]
\centering
\includegraphics[width=\textwidth]{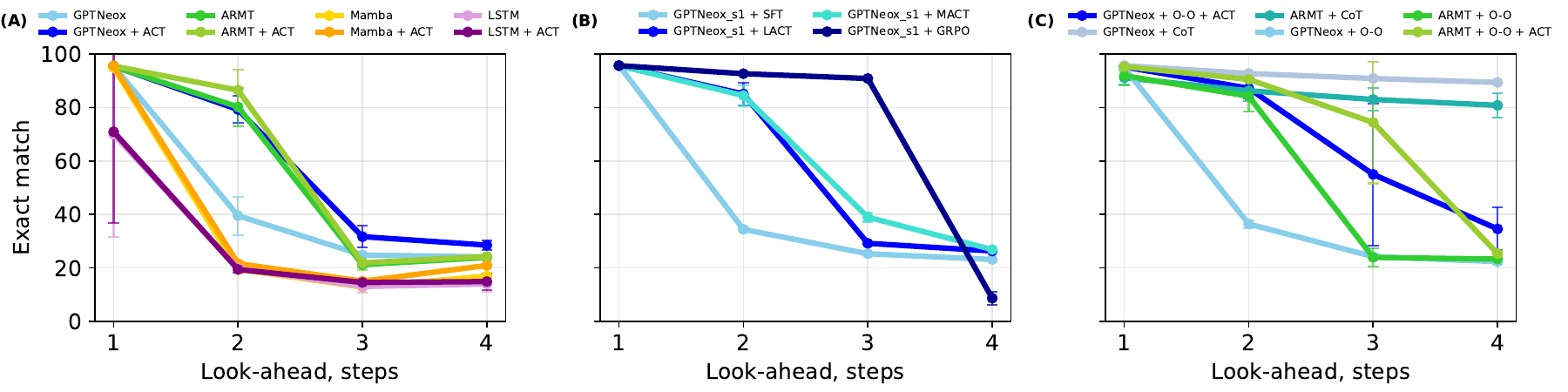}
\caption{\textbf{Extensions of computation depth enhance the  reasoning abilities of transformer-based models.} Values are exact match of the $x^{(T+k)}$ state prediction for look-ahead steps $k \in \{1, 2, 3, 4\}$. \textbf{(a)} ACT significantly improves computational abilities of transformer-based models in multi-step prediction. \textbf{(b)} Without supervision on intermediate reasoning steps RL training with GRPO allows the model to extrapolate reasoning on 3 steps forward. \textbf{(c)} With step-by-step supervision, the CoT approach significantly outperforms the in-depth approach of ACT. \gptneox and ARMT with both ACT and O-O supervision perform the best, compared to other non-CoT approaches.}
\label{linegraphs}
\end{figure*}

\section{Experiments}
\label{sec:experiments}

We start by testing contemporary LLMs on a commonsense, natural-language task that is \emph{formally equivalent} to our 1D cellular automata (1dCA) setup. The goal is to assess how well current models can (i) infer a simple logic rule from observations and (ii) chain that rule for multiple steps.

\paragraph{LLMs' performance on the \textit{Handsup} game.}
A group of friends sits around a table. In each round $n$, every friend $i$ has a binary state: \texttt{up} (hand raised) or \texttt{down}. The hidden rule has a radius $r\in\{1,2\}$: the state of friend $i$ at round $n$ depends only on the $(2r{+}1)$-tuple $\{i-r,\ldots,i,\ldots,i+r\}$ from round $n{-}1$. This is exactly a 1dCA-style local update $\rho:\{0,1\}^{2r+1}\!\to\!\{0,1\}$. 
We describe in natural language the first $T\in\{5,10\}$ rounds and ask the model to predict the behavior of players at round $T{+}k$ with $k\in\{1,2,3,4\}$. We evaluate \emph{exact-match} accuracy on the target round. To probe difficulty, we consider families of rules at $r{=}1$ and $r{=}2$, and for $r{=}1$ also group rules by Wolfram complexity class. We compare Gemini~2.5~Pro and Gemini~2.5~Flash with different “thinking budgets’’ (20k, 10k, and 0), Llama-3.3-70B, and Nemotron-32B/7B.

Figure~\ref{fig:llm-handsup} reports LLMs' performance on the Handsup game. In the simple game (Fig.~\ref{fig:llm-handsup}a) equivalent to elementary CA ($r{=}1, w{=}7, T{=}5$, 256 possible rules) only Gemini~2.5~Pro shows solid performance  with a mild decline as look-ahead increases (about $0.72\!\rightarrow\!0.69$ from $k{=}1$ to $k{=}4$). Gemini~2.5~Flash performs lower, while an extended thinking budget helps slightly over a zero budget. Llama-3.3-70B and Nemotron 7B models hover around the trivial baseline across $k$.
The hard game ($r{=}2, w{=}20, T{=}10, \approx 4.3$B possible rules) represents a strong challenge to existing LLMs (see \autoref{fig:llm-handsup}(b)), as no model crosses 10\% EM, with only Gemini models marginally above the trivial baseline. 
As Fig.~\ref{fig:llm-handsup}c shows, accuracy decreases with the dynamical complexity of the rule according to Wolfram's complexity classes. None of the models are uniformly robust across classes.


The \textit{Handsup} results directly inform our research questions. The failure of most LLMs—even with “thinking’’ budgets—to solve the simplest radius-1 setting challenges the view that current LLMs successes reflect robust generalization rather than pattern recall (RQ1). Performance degrades systematically with more complex dependencies (r=2) and higher Wolfram classes, quantifying how difficulty scales with required reasoning steps (RQ2). Finally, to disentangle whether these failures stem from architectural limits versus training/inference procedures, we proceed to controlled small-model studies that test architectural capacity under matched supervision; success would implicate training as the bottleneck, while failure would argue for architectural changes (RQ3).

\paragraph{Single-step performance across neural architectures.} We generated an 1dCA dataset with the CellPyLib~\cite{Antunes2021} for the fixed lattice size $W = 20$ and neighborhood radius $r = 2$. This configuration results in a total of $2^{2^{2r+1}} \approx 4.3\times10^9$ possible Boolean functions defining local rules. For each sample in the dataset, both the initial state and the local rule $\rho$ were generated randomly. We then computed the orbit for $T = 20$ time steps using these parameters. The training dataset consists of $9.5\times10^5$ instances and the test of $10^5$ instances. Importantly, the local rules included in the test set are exclusive and not present in the training set. 


As shown in Figure \ref{linegraphs}(a), models with different architectures can predict one step forward with nearly perfect accuracy. LSTM performs slightly worse than other architectures, likely due to challenges in effectively encoding the binary state representation. Successful learning  demonstrates that the Transformer model is capable of generalizing not only over initial conditions for a particular function — commonly the focus in studies of transformer trainability in the CA domain.
— but also across different Boolean functions of fixed arity (5 in our case).



\paragraph{Limitations in the Reasoning Depth of Transformers.} We selected a 4-layer architecture with \(d_{\text{model}}\)=128 as a baseline configuration for our experiments. Using this configuration, we separately trained from scratch for each look-ahead step $k \in \{1, 2, 3, 4\}$ of the O-S task the \gptneox \citep{black2022gptneox20bopensourceautoregressivelanguage} model to predict the state at time $x^{(T+k)}$ given an orbit $\mathcal{O}^T(x) = [x^{(1)}, x^{(2)}, \ldots, x^{(T)}]$.  

As presented in \autoref{linegraphs}(a), this task proved to be challenging. While the average accuracy for next-state prediction (O-S task with $k=1$) was 0.95, it dropped to 0.40 for $k=2$ and fell below 0.25 for $k=3$ and $k=4$. 
Despite having four layers, which in principle could capture up to two or three sequential transformations if effectively utilized, the model still struggles to learn look-ahead tasks for \(k\ge 2\). 
Specifically, the same model's depth that suffices for the single-step O-S task is no longer adequate for maintaining accurate multi-step predictions, suggesting that the capacity is being taxed by the need to encode and apply repeated rule updates in a fixed number of transformations.

To determine whether this decline was due to the \gptneox's architecture or the training objective, we explored whether accuracy could be improved by training the model to predict intermediate steps. 
This approach is analogous to multi-token prediction~\citep{gloeckle2024better}. 
We employed the Orbit-Orbit (O-O) task, training the model to predict the next four states in parallel. 
The results, also shown in \autoref{linegraphs}(c), indicate that the model's predictive abilities did not change much, though now require the curriculum learning for stable training.

This suggests that the model struggles to learn to store a hidden representation of intermediate states. Surprisingly, direct supervision of a hidden representation of the underlying rule is initially more challenging and does not facilitate better generalization to longer planning horizons (see \autoref{appx:rule_tasks}). This implies that explicitly encouraging the model to infer the generating rule cannot enhance its longer-term predictive ability by reinforcing the internalization of the system's dynamics.

The poor performance on look-ahead steps $k>1$ raises the question of whether this limitation stems from the neural network’s parameter count, layer width, or width of its embeddings. To answer this question, we performed the experiments, while varying the number of transformer layers and the embedding dimension \(d_{\text{model}}\).

Figure~\ref{fig:depth_width}\,(a) shows that accuracy for one- and two-step prediction saturates after 4–6 layers.  
Three-step prediction, however, continues to improve up to about 12 layers, whereas four-step prediction remains poor regardless of depth (up to 16 layers). \autoref{fig:depth_width}(b) examines width.  Increasing \(d_{\text{model}}\) provides only marginal gains across all horizons, with the most noticeable bump occurring between 64 and 128 dimensions; further widening yields diminishing returns. These results illustrate the importance of increasing the model's depth rather than the width of its embeddings for better multi-step reasoning performance.


\paragraph{Extending the depth of reasoning with Adaptive Computation Time.} The previous subsection confirmed that simply \emph{adding layers} offers a clear performance boost, yet even a 12-layer transformer still falters for \(k \ge 4\) (\autoref{fig:depth_width}(a)).  
 Here, we set the depth to 4 layers and study whether it is possible to improve performance by techniques that expand the \emph{effective} depth of a model at the inference time—segment-level recurrence and \textit{Adaptive Computation Time} (ACT). Hyperparameters for all models can be found in \autoref{tab:hyperparams}. Both approaches introduce additional computational steps without increasing the static layer count, potentially enabling deeper reasoning while preserving parameter efficiency.
\autoref{linegraphs}(a) shows that the auto-regressive models -- \gptneox, LSTM, and Mamba\footnote{We use the architecture from the previous section: 4 layer GPTNeox with $d_{\text{model}} = 128$ and 4 attention heads.
For Mamba, we use a state size of 16. For ARMT, $d_{\text{mem}} = 32$. As ARMT is a segment-level model, we segment our state sequence so that each segment contains a pair of consecutive states in the orbit, and the prediction is performed in the last segment using the last CA state from the input. We report the average results of 3 models trained with different seeds.}
-- handle next-state prediction but fail to solve the multi-step task. Only ARMT extends its capacity to two look-ahead steps, likely because it processes sequences segment by segment and is thus forced to separate rule and state representations. This separation may enable the generation of a hidden representation of the intermediate state, followed by the application of the rule, effectively enhancing the model's reasoning depth.


Augmenting models with ACT
has little effect on all architectures except \gptneox, which sees improved performance at \(k=2\) but not at \(k=3,4\). Overall, ARMT makes effective use of the transformer’s four-layer depth but cannot extend beyond it. Likewise, while ACT helps the transformer make use of its existing layers more efficiently, it fails to enable any architecture to solve three- or four-step predictions. Moreover, LSTM and Mamba are unable to master multi-step tasks with or without ACT, likely due to representation bottlenecks in their hidden states. 


We subsequently chose to train \gptneox model that is already capable of performing one-step reasoning with the SFT, LACT, MACT, and GRPO methods, with the goal of enabling it to reason over multiple steps without access to supervision for the intermediate reasoning stages. As illustrated in \autoref{linegraphs}(b), standard supervised fine-tuning (SFT) fails to address the problem effectively. Although the model is primarily trained on a one-step prediction task, it struggles to apply the rule iteratively. Consistent with previous results (\autoref{linegraphs}(a)), applying ACT both at the layer level (LACT) and across the entire model (MACT) improves performance on the two-step prediction task but does not generalize beyond that. Interestingly, when trained using RL (GRPO) and granted the capability to autoregressively generate intermediate ``thinking'' tokens before producing the final output, the model succeeds on the three-step prediction task. The reward signal is defined as the average token-level accuracy of the model’s prediction following the end-of-thinking token.



\paragraph{Reasoning Supervision.} We examine the impact of reasoning supervision on \gptneox and ARMT, along with their corresponding ACT-augmented variants. We replicate the O‐O training setup that incorporates mask tokens into autoregressive models within a causal masking framework. \autoref{linegraphs}(c) shows that, contrary to our expectations, the O‐O training objective alone does not yield performance improvements for either \gptneox or ARMT. However, the integration of O‐O training with ACT results in superior performance, surpassing both the baseline and ACT-only variants. To reduce variance in O-O training, we employed a curriculum learning strategy: for each $k$, training was initialized from the checkpoint obtained at $k-1$. The final reported results correspond to the checkpoint achieving the highest depth-of-reasoning score.

As a final step, we combined \gptneox and ARMT with a token‐by‐token CoT‐like next‐token prediction training. Under this regime, both models succeed at multi‐step prediction up to \(k=4\), with \gptneox slightly outperforming ARMT across each look‐ahead distance (\autoref{linegraphs}(c)). These results suggest that when explicit reasoning supervision is available, a chain-of-thought-inspired approach to training is particularly effective for enabling multi-step reasoning.
In addition to the cellular automata experiments, in Appendix \ref{appx:grmul} we show the significance of our findings on group multiplication benchmark \citep{merrill2024illusion, peng2025rwkv7gooseexpressivedynamic}. The experimental results with group multiplication show that the required depth does not increase much for recurrent models, consistent with our main findings.

\section{Discussion and Conclusions}
\label{sec:discussion}

\begin{figure*}[h]
    \centering
    \includegraphics[width=0.7\linewidth]{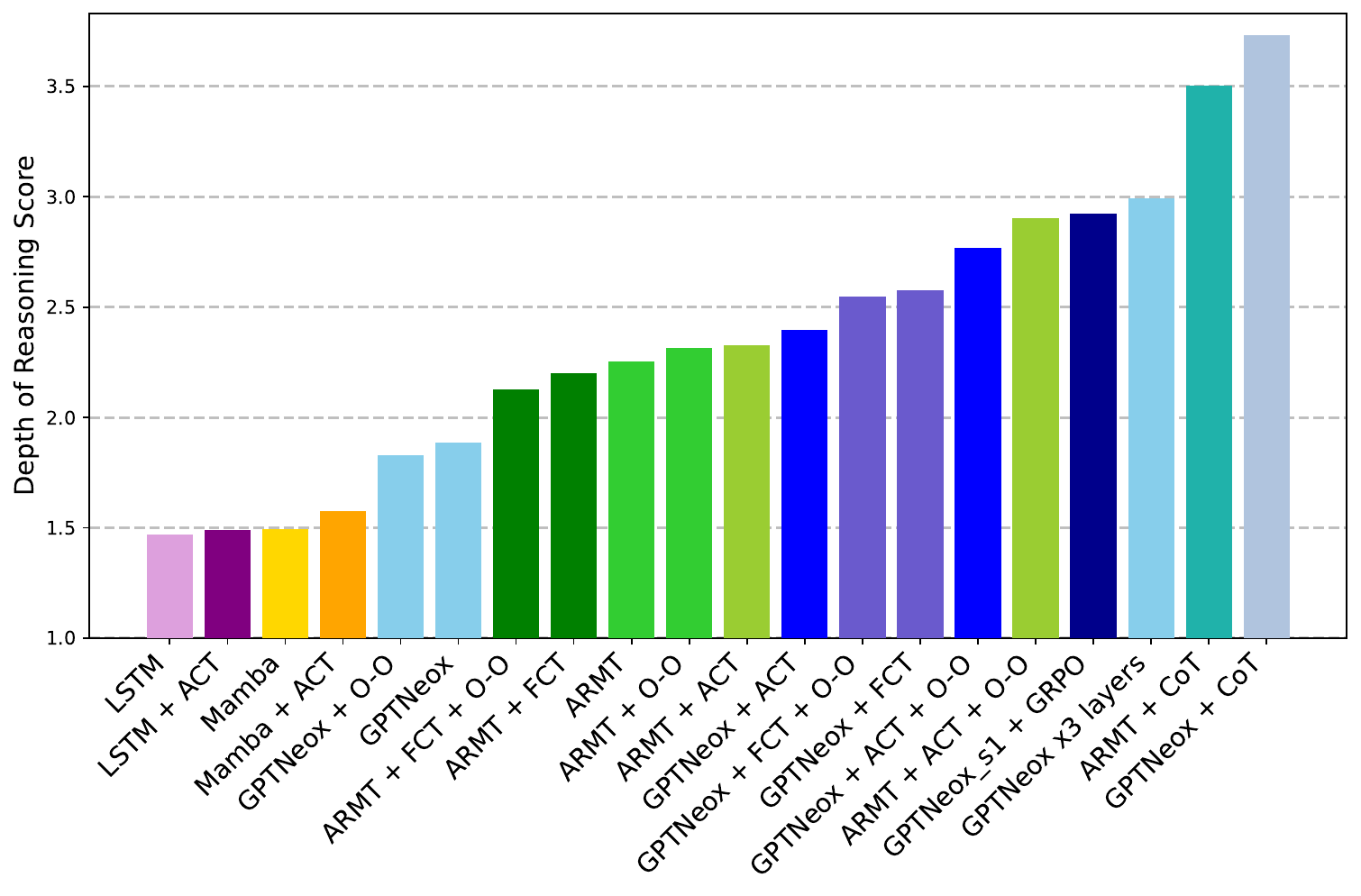}
    \caption{\textbf{With GRPO as well as with ACT and Orbit-Orbit training depth of reasoning can be significantly extended.} Average $Depth Score = 1 + \sum_{k=2}^4{\text{exact\_match}(k)}$, where $\text{exact\_match}(k)$ is the exact match accuracy of predicting the $(10+k)$th  state based on the first $10$ states. Results also include FCT (Fixed Computation Time) ablation which is described in \autoref{appx:fct}}
    \label{fig:scores}
\end{figure*}

We examine how \emph{architecture}, \emph{training signal}, and \emph{depth-extension strategy} jointly determine a model's ability to learn multi-step reasoning in 1dCA—\emph{without memorization}, since train/test rules are disjoint. The headline results (aggregated in \autoref{fig:scores}) address our \textbf{RQ1–RQ3}.

\begin{compactitem}
    \item \textbf{Models can infer unseen rules, but LLMs falter on the simplest case (\textbf{RQ1}).} Both Transformers and recurrent/SSM variants (GPT-NeoX, LSTM, Mamba, ARMT) succeed on rule induction from orbits—evidence of genuine generalization because evaluation uses unseen rules. However, evaluated LLMs (except Gemini~2.5~Pro) fail to reliably solve even the radius-1 \emph{Handsup} setting, suggesting that scale and generic “think more’’ prompting are insufficient.
    \item \textbf{Reasoning difficulty grows sharply with look-ahead depth (\textbf{RQ2}).} Fixed-depth (4-layer) models solve \(k{=}1\) but collapse for \(k{\ge}2\), revealing a clear depth barrier.
    \item \textbf{Adaptive halting adds \(\mathbf{\approx+1}\) effective step to the transformer-based architectures at low computational cost (\textbf{RQ3}).} Adding Adaptive Computation Time (ACT) to a Transformer consistently shifts the depth frontier without increasing parameters, with diminishing returns past \(k{\approx}3\).
    \item \textbf{GRPO reaches three-step rollouts without intermediate supervision (\textbf{RQ3}).} RL rewarding final correctness matches CoT performance at \(k{=}3\).
    \item \textbf{Token-level CoT saturates the current benchmark up to four steps (\textbf{RQ3}).} With stepwise targets, GPT-NeoX attains $>90\%$ accuracy for \(k{\le}4\) (\autoref{linegraphs}(c)), showing that explicit supervision can elicit deeper computation given the availability of intermediate labels.
    \item \textbf{Depth limits align with capacity constraints and can be partially mitigated (\textbf{RQ2}/\textbf{RQ3}).} Models with shallow effective depth (e.g., TC\textsuperscript{0}-like limitations) require more layers to track longer computations; ACT partially alleviates this on harder state-tracking (e.g., group-multiplication) tasks but does not fully resolve the gap.
\end{compactitem}


\paragraph{Broader implications for LLM reasoning—and beyond.}
Our results align with a growing body of evidence that \emph{reasoning failures often stem from insufficient depth allocation and sparse optimization signals}.  
For LLMs, this suggests that (i) \textbf{reasoning supervision during training is essential for reasoning}: unless intermediate steps are reinforced—via CoT, search-augmented decoding, or RL-style self-critique—models tend to default to shallow heuristics; (ii) \textbf{adaptive-depth mechanisms are a promising scaling direction}: ACT-style halting, deployed token-wise or layer-wise, can allocate computation on demand to match the variable complexity of real queries; and (iii) \textbf{explicit intermediate representations remain the most reliable route} to multi-step generalisation via CoT.
Beyond language, the same principles apply to neural algorithmic reasoning, robotic planning, and scientific simulation: whenever the target task contains latent iterative structure, giving the network \emph{room}—via dynamic recurrence, learned halting, or supervised scratch-pads—to run the hidden algorithm is more data-efficient than brute-force depth.  
We therefore advocate future benchmarks that (a) separate rule induction from state propagation, (b) report \emph{effective depth} alongside accuracy, and (c) evaluate adaptive-computation policies explicitly.  
Progress along these axes will benefit not only next-generation LLMs but also neural systems tasked with symbolic manipulation, formal verification, and open-ended planning.

\paragraph{Conclusions.}
We introduced the 1dCA reasoning benchmark that isolates multi-step reasoning \emph{without memorization} by using disjoint train/test rule sets. Success, therefore, reflects genuine \emph{rule inference} followed by iterative application, not lookup.


These findings support the main contributions of our work: (1) a benchmark that cleanly separates rule induction from state propagation; (2) a systematic comparison of architectures across this setting; (3) an analysis of depth-extension mechanisms, including recurrence, halting, reinforcement learning, and explicit stepwise supervision; and (4) practical guidance for eliciting deeper computation in a reliable and scalable way. More broadly, our results show that \emph{how} we train models and allocate computation can matter just as much as \emph{what} we train them on: training objectives that enforce multi-step prediction and mechanisms that adaptively allocate depth play a central role, while explicit intermediate representations remain the most consistent and reliable path toward deeper generalization.



\paragraph{Reproducibility.}
To facilitate reproducibility and encourage further research, we release the complete codebase, including training scripts, evaluation pipelines, and configuration files, at:
\href{https://github.com/RodkinIvan/associative-recurrent-memory-transformer/tree/ACT}{https://github.com/RodkinIvan/associative-recurrent-memory-transformer/tree/ACT}.
The repository contains all components needed to replicate the experiments, including data generation for the 1dCA benchmark and implementations of the evaluated models and training strategies.

We also provide documentation and instructions for reproducing the results. For a detailed analysis of experimental variability, including error bars and confidence intervals, see \autoref{appx:reproduce}.

\section*{Limitations}
\label{sec:limitations}

Our study uses a controlled benchmark to isolate multi-step reasoning, but this abstraction limits external validity. The setting is deterministic, fully observable, and synthetic, and may not reflect the ambiguity and complexity of real-world reasoning tasks. Most experiments are conducted on small models, and while we evaluate several LLMs via a natural-language proxy, coverage remains limited. Our use of exact-match accuracy emphasizes strict correctness but may understate partial reasoning ability, and the evaluation pipeline introduces minor extraction noise. We also explore only a subset of training and inference strategies, and success on the benchmark does not guarantee interpretable or generalizable reasoning.

\section*{Ethics and Broader Impact}

This work highlights limitations in current reasoning systems, which is important for reliable deployment in high-stakes settings. While the benchmark avoids societal biases due to its synthetic nature, these issues remain relevant when transferring insights to real-world data and applications. Improved reasoning capabilities may have dual-use risks, but our findings primarily expose constraints rather than enable immediate advances. We promote transparency and reproducibility through controlled evaluation and open code release practices.


\section*{Acknowledgments}

We would like to thank the anonymous reviewers for their constructive feedback, which has helped us improve the quality of the paper.

This work is supported by a grant \#848011 from the MBZUAI \& WIS Collaborative Research Program.

\bibliography{custom}

\begin{thebibliography}{60}
\providecommand{\natexlab}[1]{#1}

\bibitem[{Antunes(2021)}]{Antunes2021}
Luis~M. Antunes. 2021.
\newblock \href {https://doi.org/10.21105/joss.03608} {{CellPyLib}: A python library for working with cellular automata}.
\newblock \emph{Journal of Open Source Software}, 6(67):3608.

\bibitem[{Bhattamishra et~al.(2020)Bhattamishra, Patel, and Goyal}]{bhattamishra2020computational}
Satwik Bhattamishra, Arkil Patel, and Navin Goyal. 2020.
\newblock \href {https://arxiv.org/pdf/2006.09286} {On the computational power of transformers and its implications in sequence modeling}.
\newblock \emph{arXiv preprint arXiv:2006.09286}.

\bibitem[{Black et~al.(2022)Black, Biderman, Hallahan, Anthony, Gao, Golding, He, Leahy, McDonell, Phang, Pieler, Prashanth, Purohit, Reynolds, Tow, Wang, and Weinbach}]{black2022gptneox20bopensourceautoregressivelanguage}
Sid Black, Stella Biderman, Eric Hallahan, Quentin Anthony, Leo Gao, Laurence Golding, Horace He, Connor Leahy, Kyle McDonell, Jason Phang, Michael Pieler, USVSN~Sai Prashanth, Shivanshu Purohit, Laria Reynolds, Jonathan Tow, Ben Wang, and Samuel Weinbach. 2022.
\newblock \href {https://arxiv.org/abs/2204.06745} {{GPT-NeoX-20B}: An open-source autoregressive language model}.
\newblock \emph{arXiv preprint arXiv:2204.06745}.

\bibitem[{Bulatov et~al.(2022)Bulatov, Kuratov, and Burtsev}]{rmt_2022}
Aydar Bulatov, Yury Kuratov, and Mikhail Burtsev. 2022.
\newblock \href {https://proceedings.neurips.cc/paper_files/paper/2022/file/47e288629a6996a17ce50b90a056a0e1-Paper-Conference.pdf} {Recurrent memory transformer}.
\newblock In \emph{Proceedings of the 36th International Conference on Neural Information Processing Systems}, volume~35, pages 11079--11091, New Orleans, LA, USA. Curran Associates, Inc.

\bibitem[{Chevalier et~al.(2023)Chevalier, Wettig, Ajith, and Chen}]{chevalier2023adapting}
Alexis Chevalier, Alexander Wettig, Anirudh Ajith, and Danqi Chen. 2023.
\newblock \href {https://doi.org/10.18653/v1/2023.emnlp-main.232} {Adapting language models to compress contexts}.
\newblock In \emph{Proceedings of the 2023 Conference on Empirical Methods in Natural Language Processing}, pages 3829--3846, Singapore. Association for Computational Linguistics.

\bibitem[{Cybenko(1989)}]{cybenko1989approximations}
George Cybenko. 1989.
\newblock \href {https://link.springer.com/article/10.1007/BF02551274} {Approximations by superpositions of a sigmoidal function}.
\newblock \emph{Mathematics of Control, Signals and Systems}, 2:183--192.

\bibitem[{Dai et~al.(2019)Dai, Yang, Yang, Carbonell, Le, and Salakhutdinov}]{dai2019transformerxl}
Zihang Dai, Zhilin Yang, Yiming Yang, Jaime~G Carbonell, Quoc Le, and Ruslan Salakhutdinov. 2019.
\newblock \href {https://aclanthology.org/P19-1285.pdf} {Transformer-{XL}: Attentive language models beyond a fixed-length context}.
\newblock In \emph{Proceedings of the 57th Annual Meeting of the Association for Computational Linguistics}, pages 2978--2988, Florence, Italy. Association for Computational Linguistics.

\bibitem[{Dehghani et~al.(2019)Dehghani, Gouws, Vinyals, Uszkoreit, and Kaiser}]{dehghani2018universal-transformer}
Mostafa Dehghani, Stephan Gouws, Oriol Vinyals, Jakob Uszkoreit, and Lukasz Kaiser. 2019.
\newblock \href {https://openreview.net/forum?id=HyzdRiR9Y7} {Universal transformers}.
\newblock In \emph{The Seventh International Conference on Learning Representations}, New Orleans, LA, USA.

\bibitem[{Del{\'{e}}tang et~al.(2023)Del{\'{e}}tang, Ruoss, Grau{-}Moya, Genewein, Wenliang, Catt, Cundy, Hutter, Legg, Veness, and Ortega}]{deletan2023gneural}
Gr{\'{e}}goire Del{\'{e}}tang, Anian Ruoss, Jordi Grau{-}Moya, Tim Genewein, Li~Kevin Wenliang, Elliot Catt, Chris Cundy, Marcus Hutter, Shane Legg, Joel Veness, and Pedro~A. Ortega. 2023.
\newblock \href {https://openreview.net/forum?id=WbxHAzkeQcn} {Neural networks and the {C}homsky hierarchy}.
\newblock In \emph{The Eleventh International Conference on Learning Representations}, Kigali, Rwanda.

\bibitem[{Dziri et~al.(2023)Dziri, Lu, Sclar, Li, Jiang, Lin, Welleck, West, Bhagavatula, Le~Bras, Hwang, Sanyal, Ren, Ettinger, Harchaoui, and Choi}]{dziri2024faith}
Nouha Dziri, Ximing Lu, Melanie Sclar, Xiang~(Lorraine) Li, Liwei Jiang, Bill~Yuchen Lin, Sean Welleck, Peter West, Chandra Bhagavatula, Ronan Le~Bras, Jena Hwang, Soumya Sanyal, Xiang Ren, Allyson Ettinger, Zaid Harchaoui, and Yejin Choi. 2023.
\newblock \href {https://proceedings.neurips.cc/paper_files/paper/2023/file/deb3c28192f979302c157cb653c15e90-Paper-Conference.pdf} {Faith and fate: Limits of transformers on compositionality}.
\newblock In \emph{Proceedings of the 37th International Conference on Neural Information Processing Systems}, volume~36, pages 70293--70332, New Orleans, LA, USA. Curran Associates, Inc.

\bibitem[{Feng et~al.(2024)Feng, Zhang, Gu, Ye, He, and Wang}]{feng2024towards}
Guhao Feng, Bohang Zhang, Yuntian Gu, Haotian Ye, Di~He, and Liwei Wang. 2024.
\newblock \href {https://proceedings.neurips.cc/paper_files/paper/2023/hash/dfc310e81992d2e4cedc09ac47eff13e-Abstract-Conference.html} {Towards revealing the mystery behind chain of thought: a theoretical perspective}.
\newblock In \emph{Proceedings of the 37th International Conference on Neural Information Processing System}, volume~36, New Orleans, LA, USA.

\bibitem[{Gandarela et~al.(2024)Gandarela, Carvalho, and Freitas}]{gandarela2024inductive}
Jo{\~a}o~Pedro Gandarela, Danilo~S Carvalho, and Andr{\'e} Freitas. 2024.
\newblock \href {https://arxiv.org/pdf/2408.16779v1} {Inductive learning of logical theories with {LLMs}: A complexity-graded analysis}.
\newblock \emph{arXiv preprint arXiv:2408.16779}.

\bibitem[{Gloeckle et~al.(2024)Gloeckle, Idrissi, Rozi{\`e}re, Lopez-Paz, and Synnaeve}]{gloeckle2024better}
Fabian Gloeckle, Badr~Youbi Idrissi, Baptiste Rozi{\`e}re, David Lopez-Paz, and Gabriel Synnaeve. 2024.
\newblock \href {https://proceedings.mlr.press/v235/gloeckle24a.html} {Better \& faster large language models via multi-token prediction}.
\newblock In \emph{Proceedings of the 41st International Conference on Machine Learning}, Vienna, Austria. JMLR.org.

\bibitem[{Goyal et~al.(2024)Goyal, Ji, Rawat, Menon, Kumar, and Nagarajan}]{goyalthink}
Sachin Goyal, Ziwei Ji, Ankit~Singh Rawat, Aditya~Krishna Menon, Sanjiv Kumar, and Vaishnavh Nagarajan. 2024.
\newblock \href {https://openreview.net/forum?id=ph04CRkPdC} {Think before you speak: Training language models with pause tokens}.
\newblock In \emph{The Twelfth International Conference on Learning Representations}, Vienna, Austria.

\bibitem[{Graves(2016)}]{graves2016adaptive}
Alex Graves. 2016.
\newblock \href {https://arxiv.org/pdf/1603.08983} {Adaptive computation time for recurrent neural networks}.
\newblock \emph{arXiv preprint arXiv:1603.08983}.

\bibitem[{Graves et~al.(2014)Graves, Wayne, and Danihelka}]{graves2014neural}
Alex Graves, Greg Wayne, and Ivo Danihelka. 2014.
\newblock \href {https://arxiv.org/pdf/1410.5401} {Neural {T}uring machines}.
\newblock \emph{arXiv preprint arXiv:1410.5401}.

\bibitem[{Gu and Dao(2024)}]{gu2023mamba}
Albert Gu and Tri Dao. 2024.
\newblock \href {https://openreview.net/forum?id=tEYskw1VY2} {Mamba: Linear-time sequence modeling with selective state spaces}.
\newblock In \emph{First Conference on Language Modeling}, Philadelphia, PA, USA.

\bibitem[{Gu et~al.(2022)Gu, Goel, and R{\'e}}]{gu2021s4}
Albert Gu, Karan Goel, and Christopher R{\'e}. 2022.
\newblock \href {https://openreview.net/forum?id=uYLFoz1vlAC} {Efficiently modeling long sequences with structured state spaces}.
\newblock In \emph{Proceedings of the Tenth International Conference on Learning Representations}, Virtual.

\bibitem[{Guo et~al.(2025)Guo, Yang, Zhang, Song, Wang, Zhu, Xu, Zhang, Ma, Bi, Zhang, Yu, Wu, Wu, Gou, Shao, Li, Gao, Liu, Xue, Wang, Wu, Feng, Lu, Zhao, Deng, Ruan, Dai, Chen, Ji, Li, Lin, Dai, Luo, Hao, Chen, Li, Zhang, Xu, Ding, Gao, Qu, Li, Guo, Li, Chen, Yuan, Tu, Qiu, Li, Cai, Ni, Liang, Chen, Dong, Hu, You, Gao, Guan, Huang, Yu, Wang, Zhang, Zhao, Wang, Zhang, Xu, Xia, Zhang, Zhang, Tang, Zhou, Li, Wang, Li, Tian, Huang, Zhang, Wang, Chen, Du, Ge, Zhang, Pan, Wang, Chen, Jin, Chen, Lu, Zhou, Chen, Ye, Wang, Yu, Zhou, Pan, Li, Zhou, Wu, Yun, Pei, Sun, Wang, Zeng, Liu, Liang, Gao, Yu, Zhang, Xiao, An, Liu, Wang, Chen, Nie, Cheng, Liu, Xie, Liu, Yang, Li, Su, Lin, Li, Jin, Shen, Chen, Sun, Wang, Song, Zhou, Wang, Shan, Li, Wang, Wei, Zhang, Xu, Li, Zhao, Sun, Wang, Yu, Zhang, Shi, Xiong, He, Piao, Wang, Tan, Ma, Liu, Guo, Ou, Wang, Gong, Zou, He, Xiong, Luo, You, Liu, Zhou, Zhu, Huang, Li, Zheng, Zhu, Ma, Tang, Zha, Yan, Ren, Ren, Sha, Fu, Xu, Xie, Zhang, Hao, Ma, Yan, Wu, Gu, Zhu, Liu, Li, Xie, Song,
  Pan, Huang, Xu, Zhang, and Zhang}]{guo2025deepseek}
Daya Guo, Dejian Yang, Haowei Zhang, Junxiao Song, Peiyi Wang, Qihao Zhu, Runxin Xu, Ruoyu Zhang, Shirong Ma, Xiao Bi, Xiaokang Zhang, Xingkai Yu, Yu~Wu, Z.~F. Wu, Zhibin Gou, Zhihong Shao, Zhuoshu Li, Ziyi Gao, Aixin Liu, and 175 others. 2025.
\newblock \href {https://doi.org/10.1038/s41586-025-09422-z} {{DeepSeek-R1} incentivizes reasoning in {LLMs} through reinforcement learning}.
\newblock \emph{Nature}, 645(8081):633–638.

\bibitem[{Hao et~al.(2024)Hao, Sukhbaatar, Su, Li, Hu, Weston, and Tian}]{hao2024coconut}
Shibo Hao, Sainbayar Sukhbaatar, DiJia Su, Xian Li, Zhiting Hu, Jason Weston, and Yuandong Tian. 2024.
\newblock \href {https://arxiv.org/abs/2412.06769} {Training large language models to reason in a continuous latent space}.
\newblock \emph{arXiv preprint arXiv:2412.06769}.

\bibitem[{Havrilla et~al.(2024)Havrilla, Du, Raparthy, Nalmpantis, Dwivedi-Yu, Zhuravinskyi, Hambro, Sukhbaatar, and Raileanu}]{havrilla2024teaching}
Alex Havrilla, Yuqing Du, Sharath~Chandra Raparthy, Christoforos Nalmpantis, Jane Dwivedi-Yu, Maksym Zhuravinskyi, Eric Hambro, Sainbayar Sukhbaatar, and Roberta Raileanu. 2024.
\newblock \href {https://arxiv.org/pdf/2403.04642} {Teaching large language models to reason with reinforcement learning}.
\newblock \emph{arXiv preprint arXiv:2403.04642}.

\bibitem[{Herel and Mikolov(2024)}]{herel2024thinking}
David Herel and Tomas Mikolov. 2024.
\newblock \href {https://arxiv.org/pdf/2405.08644} {Thinking tokens for language modeling}.
\newblock \emph{arXiv preprint arXiv:2405.08644}.

\bibitem[{Hochreiter and Schmidhuber(1997)}]{hochreiter1997long}
Sepp Hochreiter and J{\"u}rgen Schmidhuber. 1997.
\newblock \href {https://doi.org/10.1162/neco.1997.9.8.1735} {Long short-term memory}.
\newblock \emph{Neural computation}, 9(8):1735--1780.

\bibitem[{Holliday et~al.(2024)Holliday, Mandelkern, and Zhang}]{holliday2024conditional}
Wesley~H. Holliday, Matthew Mandelkern, and Cedegao~E. Zhang. 2024.
\newblock \href {https://doi.org/10.18653/v1/2024.emnlp-main.222} {Conditional and modal reasoning in large language models}.
\newblock In \emph{Proceedings of the 2024 Conference on Empirical Methods in Natural Language Processing}, pages 3800--3821, Miami, Florida, USA. Association for Computational Linguistics.

\bibitem[{Hornik et~al.(1989)Hornik, Stinchcombe, and White}]{hornik1989multilayer}
Kurt Hornik, Maxwell Stinchcombe, and Halbert White. 1989.
\newblock \href {https://www.cs.cmu.edu/~epxing/Class/10715/reading/Kornick_et_al.pdf} {Multilayer feedforward networks are universal approximators}.
\newblock \emph{Neural networks}, 2(5):359--366.

\bibitem[{Karloff et~al.(2010)Karloff, Suri, and Vassilvitskii}]{karloff2010mpc}
Howard Karloff, Siddharth Suri, and Sergei Vassilvitskii. 2010.
\newblock \href {https://epubs.siam.org/doi/10.1137/1.9781611973075.76} {A model of computation for {MapReduce}}.
\newblock In \emph{Proceedings of the twenty-first annual ACM-SIAM symposium on Discrete Algorithms}, pages 938--948, Austin, TX, USA. SIAM.

\bibitem[{Kumar et~al.(2024)Kumar, Zhuang, Agarwal, Su, Co-Reyes, Singh, Baumli, Iqbal, Bishop, Roelofs, Zhang, McKinney, Shrivastava, Paduraru, Tucker, Precup, Behbahani, and Faust}]{kumar2024training}
Aviral Kumar, Vincent Zhuang, Rishabh Agarwal, Yi~Su, John~D Co-Reyes, Avi Singh, Kate Baumli, Shariq Iqbal, Colton Bishop, Rebecca Roelofs, Lei~M Zhang, Kay McKinney, Disha Shrivastava, Cosmin Paduraru, George Tucker, Doina Precup, Feryal Behbahani, and Aleksandra Faust. 2024.
\newblock \href {https://arxiv.org/abs/2409.12917} {Training language models to self-correct via reinforcement learning}.
\newblock \emph{arXiv preprint arXiv:2409.12917}.

\bibitem[{Luong and Lockhart(2025)}]{deepmind_gemini_deep_think_imo_2025}
Thang Luong and Edward Lockhart. 2025.
\newblock \href {https://deepmind.google/blog/advanced-version-of-gemini-with-deep-think-officially-achieves-gold-medal-standard-at-the-international-mathematical-olympiad/} {Advanced version of {G}emini with deep think officially achieves gold-medal standard at the international mathematical olympiad}.
\newblock Google DeepMind Blog.

\bibitem[{Merrill et~al.(2024)Merrill, Petty, and Sabharwal}]{merrill2024illusion}
William Merrill, Jackson Petty, and Ashish Sabharwal. 2024.
\newblock \href {https://proceedings.mlr.press/v235/merrill24a.html} {The illusion of state in state-space models}.
\newblock In \emph{Proceedings of the 41st International Conference on Machine Learning}, pages 35492--35506, Vienna, Austria.

\bibitem[{Merrill and Sabharwal(2023)}]{merrill2023parallelism}
William Merrill and Ashish Sabharwal. 2023.
\newblock \href {https://aclanthology.org/2023.tacl-1.31.pdf} {The parallelism tradeoff: Limitations of log-precision transformers}.
\newblock \emph{Transactions of the Association for Computational Linguistics}, 11:531--545.

\bibitem[{Merrill and Sabharwal(2024)}]{merrill2024expressive}
William Merrill and Ashish Sabharwal. 2024.
\newblock \href {https://openreview.net/forum?id=NjNGlPh8Wh} {The expressive power of transformers with chain of thought}.
\newblock In \emph{Proceedings of the Twelfth International Conference on Learning Representations}, ICLR~'15, Vienna, Austria.

\bibitem[{Merrill et~al.(2022)Merrill, Sabharwal, and Smith}]{merrill2022saturated}
William Merrill, Ashish Sabharwal, and Noah~A Smith. 2022.
\newblock \href {https://aclanthology.org/2022.tacl-1.49.pdf} {Saturated transformers are constant-depth threshold circuits}.
\newblock \emph{Transactions of the Association for Computational Linguistics}, 10:843--856.

\bibitem[{Mondorf and Plank(2024)}]{mondorf2024liar}
Philipp Mondorf and Barbara Plank. 2024.
\newblock \href {https://doi.org/10.18653/v1/2024.emnlp-main.404} {Liar, liar, logical mire: A benchmark for suppositional reasoning in large language models}.
\newblock In \emph{Proceedings of the 2024 Conference on Empirical Methods in Natural Language Processing}, pages 7114--7137, Miami, Florida, USA. Association for Computational Linguistics.

\bibitem[{Nowak et~al.(2024)Nowak, Svete, Butoi, and Cotterell}]{nowak-etal-2024-representational}
Franz Nowak, Anej Svete, Alexandra Butoi, and Ryan Cotterell. 2024.
\newblock \href {https://aclanthology.org/2024.acl-long.676} {On the representational capacity of neural language models with chain-of-thought reasoning}.
\newblock In \emph{Proceedings of the 62nd Annual Meeting of the Association for Computational Linguistics (Volume 1: Long Papers)}, pages 12510--12548, Bangkok, Thailand. Association for Computational Linguistics.

\bibitem[{{OpenAI}(2024)}]{openai2024learning}
{OpenAI}. 2024.
\newblock Learning to reason with {LLMs}.
\newblock \url{https://openai.com/index/learning-to-reason-with-llms/}.
\newblock Accessed: 2024-09-23.

\bibitem[{Ouyang et~al.(2022)Ouyang, Wu, Jiang, Almeida, Wainwright, Mishkin, Zhang, Agarwal, Slama, Ray, Schulman, Hilton, Kelton, Miller, Simens, Askell, Welinder, Christiano, Leike, and Lowe}]{ouyang2022training}
Long Ouyang, Jeffrey Wu, Xu~Jiang, Diogo Almeida, Carroll Wainwright, Pamela Mishkin, Chong Zhang, Sandhini Agarwal, Katarina Slama, Alex Ray, John Schulman, Jacob Hilton, Fraser Kelton, Luke Miller, Maddie Simens, Amanda Askell, Peter Welinder, Paul~F Christiano, Jan Leike, and Ryan Lowe. 2022.
\newblock \href {https://proceedings.neurips.cc/paper_files/paper/2022/file/b1efde53be364a73914f58805a001731-Paper-Conference.pdf} {Training language models to follow instructions with human feedback}.
\newblock In \emph{Advances in Neural Information Processing Systems}, volume~35, pages 27730--27744, New Orleans, LA, USA. Curran Associates, Inc.

\bibitem[{Peng et~al.(2025)Peng, Zhang, Goldstein, Alcaide, Du, Hou, Lin, Liu, Lu, Merrill, Song, Tan, Utpala, Wilce, Wind, Wu, Wuttke, and Zhou-Zheng}]{peng2025rwkv7gooseexpressivedynamic}
Bo~Peng, Ruichong Zhang, Daniel Goldstein, Eric Alcaide, Xingjian Du, Haowen Hou, Jiaju Lin, Jiaxing Liu, Janna Lu, William Merrill, Guangyu Song, Kaifeng Tan, Saiteja Utpala, Nathan Wilce, Johan~S. Wind, Tianyi Wu, Daniel Wuttke, and Christian Zhou-Zheng. 2025.
\newblock \href {https://arxiv.org/abs/2503.14456} {Rwkv-7 "goose" with expressive dynamic state evolution}.
\newblock \emph{Preprint}, arXiv:2503.14456.

\bibitem[{P{\'e}rez et~al.(2021)P{\'e}rez, Barcel{\'o}, and Marinkovic}]{perez2021attention}
Jorge P{\'e}rez, Pablo Barcel{\'o}, and Javier Marinkovic. 2021.
\newblock \href {https://jmlr.org/papers/volume22/20-302/20-302.pdf} {Attention is {T}uring-complete}.
\newblock \emph{Journal of Machine Learning Research}, 22(75):1--35.

\bibitem[{Pfau et~al.(2024)Pfau, Merrill, and Bowman}]{pfau2024lets}
Jacob Pfau, William Merrill, and Samuel~R. Bowman. 2024.
\newblock \href {https://openreview.net/forum?id=NikbrdtYvG} {Let{\textquoteright}s think dot by dot: Hidden computation in transformer language models}.
\newblock In \emph{First Conference on Language Modeling}, Philadelphia, PA, USA.

\bibitem[{Rae et~al.(2020)Rae, Potapenko, Jayakumar, Hillier, and Lillicrap}]{raecompressive2019}
Jack~W. Rae, Anna Potapenko, Siddhant~M. Jayakumar, Chloe Hillier, and Timothy~P. Lillicrap. 2020.
\newblock \href {https://openreview.net/forum?id=SylKikSYDH} {Compressive transformers for long-range sequence modelling}.
\newblock In \emph{The Eighth International Conference on Learning Representations}, Addis Ababa, Ethiopia.

\bibitem[{Rodkin et~al.(2024)Rodkin, Kuratov, Bulatov, and Burtsev}]{armt}
Ivan Rodkin, Yurii Kuratov, Aydar Bulatov, and Mikhail Burtsev. 2024.
\newblock \href {https://arxiv.org/abs/2407.04841} {Associative recurrent memory transformer}.
\newblock \emph{Preprint}, arXiv:2407.04841.

\bibitem[{Sanford et~al.(2024{\natexlab{a}})Sanford, Hsu, and Telgarsky}]{sanford2024transformers}
Clayton Sanford, Daniel Hsu, and Matus Telgarsky. 2024{\natexlab{a}}.
\newblock \href {https://proceedings.mlr.press/v235/sanford24a.html} {Transformers, parallel computation, and logarithmic depth}.
\newblock In \emph{Proceedings of the 41st International Conference on Machine Learning}, Vienna, Austria. JMLR.org.

\bibitem[{Sanford et~al.(2024{\natexlab{b}})Sanford, Hsu, and Telgarsky}]{sanford2024representational}
Clayton Sanford, Daniel~J Hsu, and Matus Telgarsky. 2024{\natexlab{b}}.
\newblock \href {https://proceedings.neurips.cc/paper_files/paper/2023/file/73bf692447f174984f30499ec9b20e04-Supplemental-Conference.pdf} {Representational strengths and limitations of transformers}.
\newblock In \emph{Proceedings of the 37th International Conference on Neural Information Processing Systems}, volume~36, New Orleans, LA, USA. Curran Associates Inc.

\bibitem[{Schulman et~al.(2017)Schulman, Wolski, Dhariwal, Radford, and Klimov}]{schulman2017proximal}
John Schulman, Filip Wolski, Prafulla Dhariwal, Alec Radford, and Oleg Klimov. 2017.
\newblock \href {https://arxiv.org/pdf/1707.06347} {Proximal policy optimization algorithms}.
\newblock \emph{arXiv preprint arXiv:1707.06347}.

\bibitem[{Shao et~al.(2024)Shao, Wang, Zhu, Xu, Song, Bi, Zhang, Zhang, Li, Wu, and Guo}]{shao2024deepseekmath}
Zhihong Shao, Peiyi Wang, Qihao Zhu, Runxin Xu, Junxiao Song, Xiao Bi, Haowei Zhang, Mingchuan Zhang, Y.~K. Li, Y.~Wu, and Daya Guo. 2024.
\newblock \href {https://arxiv.org/abs/2402.03300} {{DeepSeekMath}: Pushing the limits of mathematical reasoning in open language models}.
\newblock \emph{arXiv preprint arXiv:2402.03300}.

\bibitem[{Shojaee et~al.(2025)Shojaee, Mirzadeh, Alizadeh, Horton, Bengio, and Farajtabar}]{shojaee2025illusionthinkingunderstandingstrengths}
Parshin Shojaee, Iman Mirzadeh, Keivan Alizadeh, Maxwell Horton, Samy Bengio, and Mehrdad Farajtabar. 2025.
\newblock \href {https://arxiv.org/abs/2506.06941} {The illusion of thinking: Understanding the strengths and limitations of reasoning models via the lens of problem complexity}.
\newblock \emph{arXiv preprint arXiv:2506.06941}.

\bibitem[{Strobl et~al.(2024)Strobl, Merrill, Weiss, Chiang, and Angluin}]{strobl2024formal}
Lena Strobl, William Merrill, Gail Weiss, David Chiang, and Dana Angluin. 2024.
\newblock \href {https://aclanthology.org/2024.tacl-1.30.pdf} {What formal languages can transformers express? {A} survey}.
\newblock \emph{Transactions of the Association for Computational Linguistics}, 12:543--561.

\bibitem[{Uesato et~al.(2022)Uesato, Kushman, Kumar, Song, Siegel, Wang, Creswell, Irving, and Higgins}]{uesato2022solving}
Jonathan Uesato, Nate Kushman, Ramana Kumar, Francis Song, Noah Siegel, Lisa Wang, Antonia Creswell, Geoffrey Irving, and Irina Higgins. 2022.
\newblock \href {https://arxiv.org/pdf/2211.14275} {Solving math word problems with process-and outcome-based feedback}.
\newblock \emph{arXiv preprint arXiv:2211.14275}.

\bibitem[{Valmeekam et~al.(2024)Valmeekam, Stechly, and Kambhampati}]{valmeekam2024llms}
Karthik Valmeekam, Kaya Stechly, and Subbarao Kambhampati. 2024.
\newblock \href {https://arxiv.org/abs/2409.13373} {{LLMs} still can't plan; can {LRMs}? {A} preliminary evaluation of {OpenAI}'s o1 on {PlanBench}}.
\newblock \emph{arXiv preprint arXiv:2409.13373}.

\bibitem[{Vaswani et~al.(2017)Vaswani, Shazeer, Parmar, Uszkoreit, Jones, Gomez, Kaiser, and Polosukhin}]{vaswani2017attention}
Ashish Vaswani, Noam Shazeer, Niki Parmar, Jakob Uszkoreit, Llion Jones, Aidan~N Gomez, {\L}ukasz Kaiser, and Illia Polosukhin. 2017.
\newblock \href {http://papers.nips.cc/paper/7181-attention-is-all-you-need} {Attention is all you need}.
\newblock In \emph{Proceedings of the 31st International Conference on Neural Information Processing Systems}, pages 5998--6008, Long Beach, CA, USA.

\bibitem[{Wan et~al.(2024)Wan, Wang, Yang, Yuan, Huang, He, Jiao, and Lyu}]{wan2024b}
Yuxuan Wan, Wenxuan Wang, Yiliu Yang, Youliang Yuan, Jen-tse Huang, Pinjia He, Wenxiang Jiao, and Michael Lyu. 2024.
\newblock \href {https://doi.org/10.18653/v1/2024.emnlp-main.128} {{L}ogic{A}sker: Evaluating and improving the logical reasoning ability of large language models}.
\newblock In \emph{Proceedings of the 2024 Conference on Empirical Methods in Natural Language Processing}, pages 2124--2155, Miami, Florida, USA. Association for Computational Linguistics.

\bibitem[{Wang et~al.(2024)Wang, Li, Shao, Xu, Dai, Li, Chen, Wu, and Sui}]{wang2023math}
Peiyi Wang, Lei Li, Zhihong Shao, Runxin Xu, Damai Dai, Yifei Li, Deli Chen, Yu~Wu, and Zhifang Sui. 2024.
\newblock \href {https://doi.org/10.18653/v1/2024.acl-long.510} {Math-shepherd: Verify and reinforce {LLM}s step-by-step without human annotations}.
\newblock In \emph{Proceedings of the 62nd Annual Meeting of the Association for Computational Linguistics (Volume 1: Long Papers)}, pages 9426--9439, Bangkok, Thailand. Association for Computational Linguistics.

\bibitem[{Wei et~al.(2022)Wei, Wang, Schuurmans, Bosma, ichter, Xia, Chi, Le, and Zhou}]{wei2022chain}
Jason Wei, Xuezhi Wang, Dale Schuurmans, Maarten Bosma, brian ichter, Fei Xia, Ed~Chi, Quoc~V Le, and Denny Zhou. 2022.
\newblock \href {https://proceedings.neurips.cc/paper_files/paper/2022/file/9d5609613524ecf4f15af0f7b31abca4-Paper-Conference.pdf} {Chain-of-thought prompting elicits reasoning in large language models}.
\newblock In \emph{Proceedings of the 36th International Conference on Neural Information Processing Systems}, volume~35, pages 24824--24837, New Orleans, LA, USA. Curran Associates, Inc.

\bibitem[{Weston et~al.(2015)Weston, Chopra, and Bordes}]{weston2014memory}
Jason Weston, Sumit Chopra, and Antoine Bordes. 2015.
\newblock \href {http://arxiv.org/abs/1410.3916} {Memory networks}.
\newblock In \emph{Proceedings of the Third International Conference on Learning Representations}, San Diego, CA, USA.

\bibitem[{Yang et~al.(2024)Yang, Lee, Nowak, and Papailiopoulos}]{yang2024loopedtransformersbetterlearning}
Liu Yang, Kangwook Lee, Robert Nowak, and Dimitris Papailiopoulos. 2024.
\newblock \href {https://arxiv.org/abs/2311.12424} {Looped transformers are better at learning learning algorithms}.
\newblock \emph{Preprint}, arXiv:2311.12424.

\bibitem[{Yao et~al.(2024)Yao, Yu, Zhao, Shafran, Griffiths, Cao, and Narasimhan}]{yao2024tree}
Shunyu Yao, Dian Yu, Jeffrey Zhao, Izhak Shafran, Tom Griffiths, Yuan Cao, and Karthik Narasimhan. 2024.
\newblock \href {https://proceedings.neurips.cc/paper_files/paper/2023/file/271db9922b8d1f4dd7aaef84ed5ac703-Paper-Conference.pdf} {Tree of thoughts: Deliberate problem solving with large language models}.
\newblock In \emph{Proceedings of the 37th International Conference on Neural Information Processing Systems}, volume~36, New Orleans, LA, USA. Curran Associates Inc.

\bibitem[{Yu et~al.(2025)Yu, He, Li, Zhou, Zhang, Xu, and He}]{yu2025enhancingautoregressivechainofthoughtloopaligned}
Qifan Yu, Zhenyu He, Sijie Li, Xun Zhou, Jun Zhang, Jingjing Xu, and Di~He. 2025.
\newblock \href {https://arxiv.org/abs/2502.08482} {Enhancing auto-regressive chain-of-thought through loop-aligned reasoning}.
\newblock \emph{Preprint}, arXiv:2502.08482.

\bibitem[{Yun et~al.(2020)Yun, Bhojanapalli, Rawat, Reddi, and Kumar}]{yun2019transformers}
Chulhee Yun, Srinadh Bhojanapalli, Ankit~Singh Rawat, Sashank~J Reddi, and Sanjiv Kumar. 2020.
\newblock \href {https://openreview.net/forum?id=ByxRM0Ntvr} {Are transformers universal approximators of sequence-to-sequence functions?}
\newblock In \emph{Proceedings of the 8th International Conference on Learning Representations}.

\bibitem[{Zhang et~al.(2024)Zhang, Abdul-Mageed, and Lakshmanan}]{zhang2024autoregressive}
Xiang Zhang, Muhammad Abdul-Mageed, and Laks~VS Lakshmanan. 2024.
\newblock \href {https://arxiv.org/pdf/2409.09239} {Autoregressive + chain of thought = recurrent: Recurrence's role in language models' computability and a revisit of recurrent transformer}.
\newblock \emph{arXiv preprint arXiv:2409.09239}.

\bibitem[{Zhu et~al.(2025)Zhu, Wang, Hua, Zhang, Li, Que, Wei, Wen, Yin, Xing, Li, Shi, Ma, Li, Kergan, Smith, Qu, Hui, Wu, Min, Huang, Zhou, Ye, Liu, Yang, Shi, Lin, Zhao, Cai, Zhang, Huang, Bengio, and Eshraghian}]{zhu2025scalinglatentreasoninglooped}
Rui-Jie Zhu, Zixuan Wang, Kai Hua, Tianyu Zhang, Ziniu Li, Haoran Que, Boyi Wei, Zixin Wen, Fan Yin, He~Xing, Lu~Li, Jiajun Shi, Kaijing Ma, Shanda Li, Taylor Kergan, Andrew Smith, Xingwei Qu, Mude Hui, Bohong Wu, and 14 others. 2025.
\newblock \href {https://arxiv.org/abs/2510.25741} {Scaling latent reasoning via looped language models}.
\newblock \emph{Preprint}, arXiv:2510.25741.

\end{thebibliography}

\newpage
\appendix


\section{Related Work}

\textbf{Computational Expressivity.} \citet{sanford2024representational} show that in setups where the input context length grows but the model depth remains constant, transformers achieve logarithmic complexity scaling in input size for sparse averaging tasks and linear scaling for triple detection.
They further use the simulation of transformers in a constant number of MPC~\citep{karloff2010mpc} communication rounds to demonstrate their expressive power, showing that logarithmic-depth transformers can efficiently solve tasks that are intractable for graph neural networks and recurrent models~\cite{sanford2024transformers}.
\citet{merrill2023parallelism} prove that transformers with logarithmic precision can be simulated by constant-depth logspace-uniform threshold circuits, implying fundamental computational limitations. \citet{zhang2024autoregressive} employ circuit complexity theory to show that bounded-depth transformers cannot directly solve certain arithmetic or equation tasks,
unless the model size increases exponentially.

\textbf{Formal Language Recognition.} The Chomsky hierarchy has been used to classify the computational capabilities of transformers and their expressivity limits. \citet{deletan2023gneural} show that transformers struggle with non-regular languages. \citet{strobl2024formal} provide a comprehensive survey on how transformers relate to formal language classes, identifying the architectural constraints that limit their ability to process hierarchical structures. They show that while transformers with softmax attention can count, they remain within $\mathsf{TC}^0$ and struggle with evaluating Boolean formulas or solving complex hierarchical tasks.
\citet{zhang2024autoregressive} discuss transformers' limitations due to their lack of recurrence, arguing that they are computationally weaker than recurrent models in formal language tasks.

Several studies explore how CoT enhances transformer reasoning capabilities. \citet{feng2024towards} show that transformers can solve arithmetic and dynamic programming tasks via CoT, which they fail to do directly. \citet{merrill2024expressive} demonstrates that CoT increases computational power, enabling the recognition of regular languages. \citet{nowak-etal-2024-representational} formalize CoT reasoning, showing equivalence to probabilistic Turing machines. \citet{zhang2024autoregressive} argue that CoT can approximate recurrent computation, mitigating transformers'
lack of explicit recurrence.

There are generalizations of CoT that relax the human-like word-by-word out-loud reasoning. The reasoning process has been moved to special pause~\citep{goyalthink}, think~\citep{herel2024thinking}, or filler~\citep{pfau2024lets} tokens to allow the model to think internally before generating a response. Coconut (Chain of Continuous Thought)~\cite{hao2024coconut} further extends this by replacing explicit word decoding with the model's last hidden state as input to the next step, effectively shifting reasoning into the latent space.
Moreover, since real-world datasets rarely include supervision for long, multi-step reasoning, approaches that incorporate verifiers or intermediate feedback have become increasingly important~\citep{pfau2024lets}. At the same time, reinforcement learning methods~\citep{schulman2017proximal}, such as GRPO~\citep{shao2024deepseekmath}, which rely solely on rewards for correct final answers, show great promise.


Overall, these studies highlight the limitations of transformers in reasoning depth and computational power, showing that CoT-like approaches and recurrence can help mitigate these constraints.
Our work explores the use of One-dimensional Cellular Automata (1dCA) as a framework to evaluate models' reasoning abilities. 1dCA provides a flexible and controlled setting where the number of sequential steps required to solve a task can be precisely defined. Adjusting the complexity of state transition rules allows for varying task difficulty.

\paragraph{Looped Transformers} \citep{yang2024loopedtransformersbetterlearning} investigates whether looped transformers can emulate iterative learning algorithms, such as gradient descent, for data-fitting problems like linear regression. They show that looped transformers can achieve performance comparable to standard transformers with fewer parameters by replicating these iterative optimization steps. Our paper investigates how different architectures and training methods affect a model's ability to learn and perform multi-step reasoning and rule abstraction. The "iterations" in our study are interpreted as steps for applying a rule or propagating a state, which is distinct from emulating optimization algorithms.

RELAY \citep{yu2025enhancingautoregressivechainofthoughtloopaligned} aligns CoT steps with loop iterations and uses intermediate supervision in looped transformer training to generate high-quality reasoning chains for auto-regressive models. It aims to leverage looped transformers' length generalization to improve the handling of longer reasoning chains by auto-regressive models.

We study CoT as a training objective that provides direct reasoning supervision on intermediate states for multi-step state prediction on 1dCA. While both studies involve recurrence and CoT-like supervision, Yu et al.'s work focuses on a specific methodology for generating CoT for other models by aligning CoT steps with loops, whereas our work directly evaluates how training with or without intermediate supervision, as in O-O or GRPO, respectively, influences a model's core reasoning capabilities in a disentangled environment.

In the "Illusion of Thinking" research \citep{shojaee2025illusionthinkingunderstandingstrengths} authors show that the models' performance decreases with the increased complexity of puzzle environments. For thinking models, however, this degradation is less dramatic. This is consistent with our findings in \autoref{linegraphs} (a).

\section{Models Discussion}
\label{appx:models_disc}

\paragraph{LSTM}
By integrating a gating mechanism into recurrent neural networks, LSTMs alleviated the vanishing gradient problem, allowing the model to retain information from up to 10–15 prior time steps. However, LSTMs still face several limitations. First, despite the gating mechanism, they often struggle with very long-range dependencies, as information can decay over extended sequences. Second, their sequential nature hinders parallelization, which slows training and increases computational costs compared to more modern architectures such as transformers. As a result, while LSTMs represented a major breakthrough in sequence modeling and, in theory, can process contexts of infinite length, they have been largely superseded by more scalable and efficient models.

\paragraph{Transformers}
The attention mechanism allows transformers to focus on relevant parts of the input, facilitating information integration across long distances. As a result, they maintain and reuse context more effectively than LSTMs, making them a backbone for modern large language models. This design has enabled state-of-the-art performance on complex reasoning tasks, cementing the transformer's role in natural language processing. While this flexibility is powerful, it also introduces drawbacks. Transformers must compute and store a large attention matrix, often scaling to $O(n^2)$ in memory and computation. This creates challenges when handling very long inputs or generating lengthy outputs, as hardware and software limitations cap the context window.

Another limitation of transformers is their difficulty in processing information ``in-depth.'' Each generation step requires a fixed amount of computation, constrained by the number of transformer layers. Consequently, transformers face challenges with multi-hop reasoning. To enable more efficient in-depth reasoning, various test-time compute strategies have been introduced, including chain-of-thought prompting, Monte Carlo Tree Search, and others. While these techniques partially mitigate the issue, they remain bottlenecks: longer generations demand substantial computational resources and may exceed the effective context window. These techniques also require  supervision for intermediate steps to train the model. This is a huge limitation as strong AGI systems should automatically learn to recursively apply rules to data.

\paragraph{State Space Models}
While less prevalent compared to RNNs and transformers, SSMs are widely used in control theory and signal processing. In the context of neural networks, SSMs aim to combine the strengths of recurrent models, such as handling infinitely large contexts, with the efficiency of convolutional models for fast prompt processing and training. This positions SSMs as a middle ground between classical LSTMs and transformers.

In our experiments, we utilize Mamba, an SSM variant improved with a selective mechanism~\citep{gu2023mamba, gu2021s4}.
The Mamba Selective State Model extends this framework by making \( A \), \( B \), and \( C \) dynamic, adjusting them based on the input \( x(t) \). This adaptive mechanism allows Mamba to selectively focus on relevant input features, filtering out irrelevant details~\citep{gu2023mamba}.
By dynamically adapting its parameters, Mamba is able to capture long-range dependencies in sequences while remaining computationally efficient. 

While SSMs excel at efficiently modeling long-range dependencies and processing sequential data with lower computational overhead than transformers, they typically lack the expressiveness and flexibility needed for advanced reasoning tasks, and multi-step inference. In particular, these models may struggle to capture complex hierarchical relationships and to maintain rich intermediate representations over extended reasoning processes, which further compounds the limitations already present in transformers when tackling in-depth reasoning problems.

\paragraph{Associative Recurrent Memory Transformer}
As shown by \citet{armt}, ARMT can leverage information from the distant past, up to 50 million tokens. Compared to SSMs, ARMT is more expressive due to its grounding in the classical transformer architecture and introduces the ability to recurrently process contexts of infinite length.

\begin{table*}[]
    \centering
    \begin{tabular}{|c|c|c|c|c|c|}
     Model & Depth & $d_\text{model} $ & $d_\text{mem}$ / state\_size & $n_{\text{heads}}$& max ACT iterations\\
     \hline
    \gptneox & 4 & 128 & - & 4 & 4\\
    ARMT & 4 & 128 & 32 & 4 & 4\\
    Mamba & 4 & 128 & 16 & - & 4\\
    LSTM & 4 & 128 & - & - & 4 \\

\end{tabular}
    \caption{\textbf{Hyperparameters for the base models.} We used these hyperparameters in the O-S, O-O, O-RS and RO-S experiments, as well as CoT and GRPO experiments.}
    \label{tab:hyperparams}
\end{table*}

\paragraph{Theoretical Depth Estimates} Theoretical estimates predict that for GPTNeox and Mamba, the depth of computation is limited by the number of layers $Depth = O(L)$, where $L$ is the number of model layers. For LSTM, computational depth not only grows with the number of layers but also with the sequence length, making $Depth = O(L+N)$; here $N$ is the sequence length. ARMT is a trade-off between parallelization and recurrence. It utilizes the forward transformer for local processing of the segment but passes its recurrent state between segments in an RNN-like format, which allows its computational depth to grow with the sequence length, making $Depth = O(L + \frac{N}{S})$; here $S$ is the segment size.

\section{Reproducibility Statement}
\label{appx:reproduce}

Metrics are reported with 95\% confidence intervals for the Handsup game with language models. In all small model finetuning experiments, we report standard deviation estimates (the square root of unbiased variance estimation) for confidence intervals. All hyperparameters are specified in \autoref{tab:hyperparams}, and we describe training details and used hardware in Section~\ref{sec:training_details}. We also release the full codebase to ensure the reproducibility of results. It can be found on our GitHub: \href{https://github.com/RodkinIvan/associative-recurrent-memory-transformer/tree/ACT}{https://github.com/RodkinIvan/associative-recurrent-memory-transformer/tree/ACT}.

\section{Training Details}
\label{sec:training_details}

We train all models for 40k optimization steps using the Adam optimizer with a learning rate of 3e-4, together with a linear warmup schedule over the first 1000 steps followed by linear decay for the remainder of training. We use a total batch size of 256 samples across all experiments. The vast majority of experiments were conducted on a single NVIDIA RTX 6000 Ada GPU. The full set of model hyperparameters is provided in \autoref{tab:hyperparams}.

\section{Adaptive Computation Time Formulation}
\label{appx:act}
The module calculates a halting weight $p_t$ at each computation step $t$, which represents the percentage of the task completed by the module $f$:
    \begin{gather}
        p_t = \text{HALT}(h_t), \quad
        h_{t+1} = f(h_t), \\
        \text{HALT}(h_t) = \sigma(W_h h_t + b_h)
    \end{gather} 
    where $h_t$ is the layer input.
This weight is accumulated  into $P_t$ until the halting condition is met:
    \begin{gather} \textstyle
        P_t = \sum_{i=0}^t p_i, \\
        T = \text{argmin}_t(P_t \ge 1 - \epsilon) + 1.
    \end{gather} 
    Finally, the prediction is done in the following way:
 $ y = \sum_{t=0}^{T-1} p_t h_{t+1} $ with $ p_{T-1} = R = 1 - \sum_{t=0} ^{T-2} p_t$. For training, we add an auxiliary component to the loss function $\hat{L} = L +\tau R$. This component serves as a time penalty.

\section{Rule-based Task Variants}
\label{appx:rule_tasks}
\begin{figure*}[t]
\centering
\subfigure[State prediction]{\includegraphics[width=0.25\textwidth]{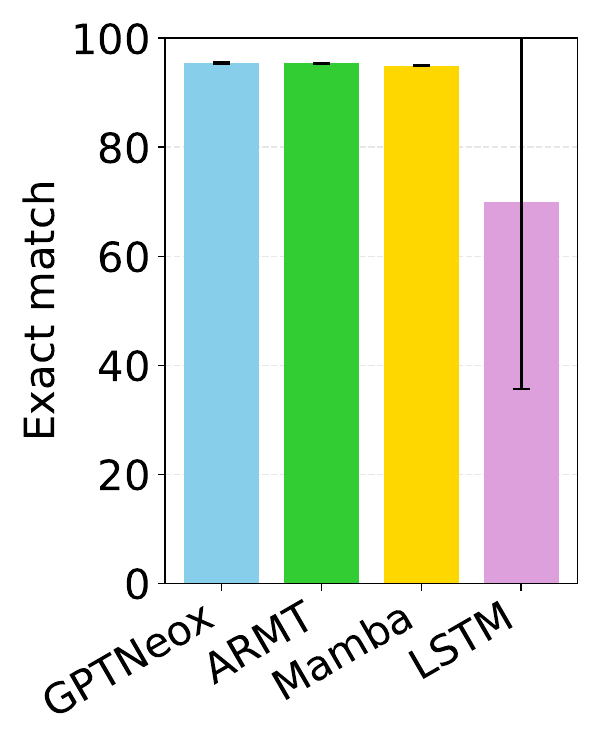}\label{fig:o-s_s1}}
\hfill
\subfigure[Rule prediction]{\includegraphics[width=0.25\textwidth]{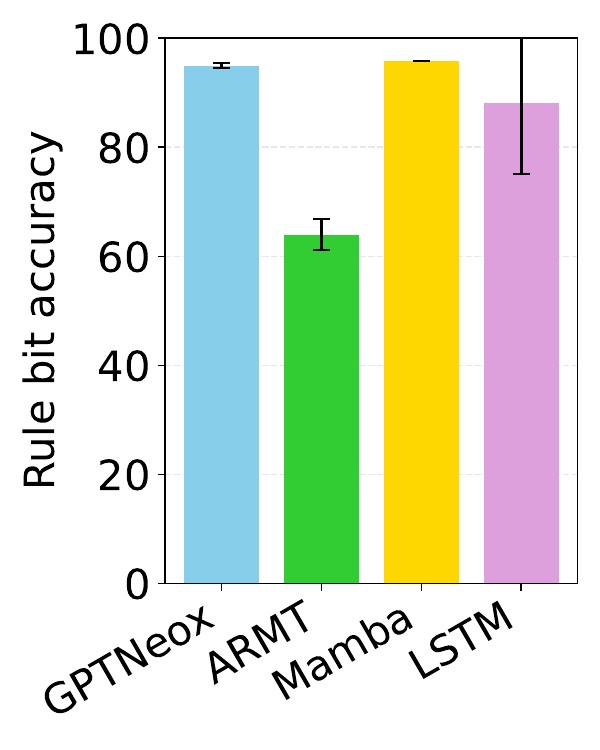}\label{fig:o-rs_s1}}
\hfill
\subfigure[\gptneox on O-S, O-O, O-RS, RO-S tasks]{\includegraphics[width=0.48\textwidth]{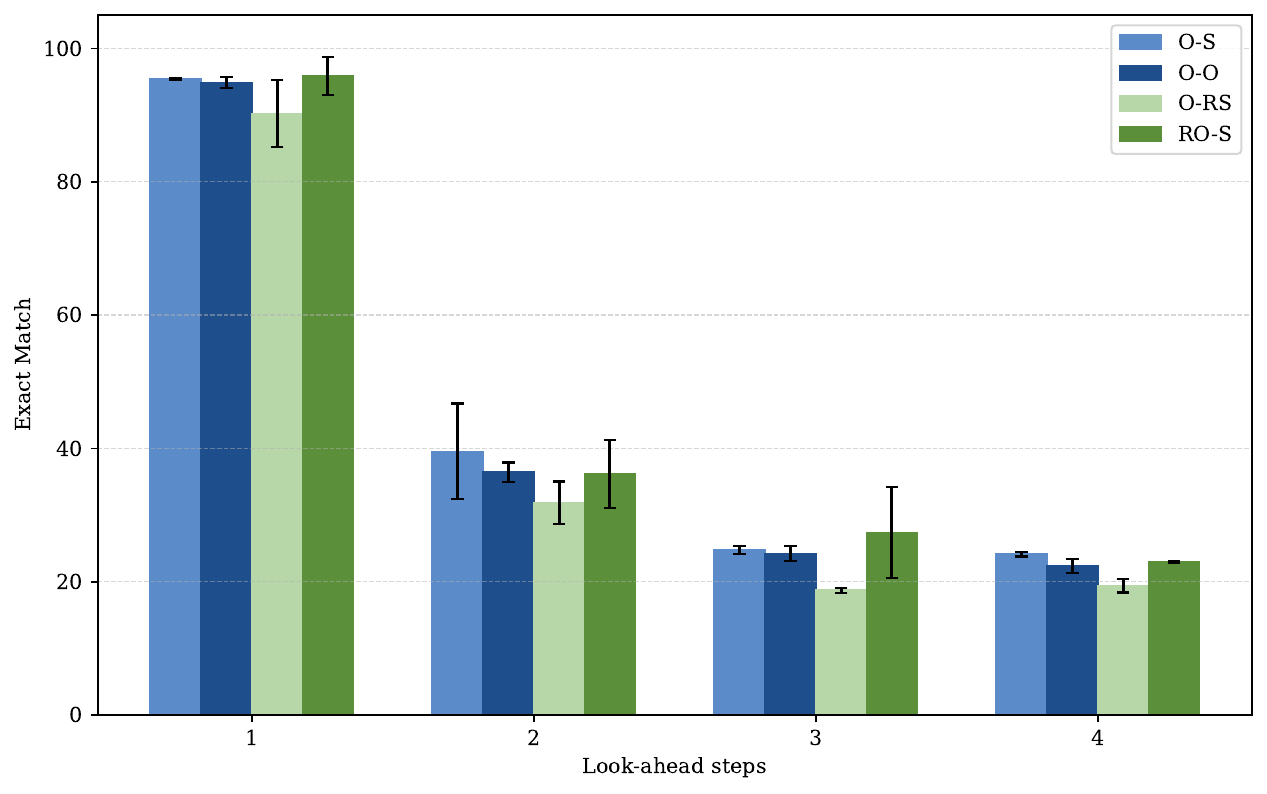}\label{fig:gptneox_combined}}
\caption{\textbf{Single-step accuracy is near-perfect across models, but multi-step performance collapses.}
\textbf{(a)}~Exact-match accuracy for single-step \emph{state prediction} (O-S): all models except LSTM achieve >95 \%.  
\textbf{(b)}~Bit-wise accuracy for \emph{rule inference} (O-RS): most architectures recover the hidden Boolean rule, yet ARMT trails the rest.  
\textbf{(c)}~\gptneox\ accuracy on variable-horizon prediction across the four task variants (O-S, O-O, O-RS, RO-S): accuracy falls steeply with look-ahead \(k\). }
\label{fig:combined_performance}
\end{figure*}

 We additionally consider two variations of learning tasks designed to assess different aspects of predictive modeling and rule inference:

\textit{Orbit-State and Rule (O-RS)}: given an orbit $\mathcal{O}^T(x)$ predict the state $x^{(T+k)}$ and the local rule $\rho$. By explicitly optimizing rule prediction, the model receives direct supervision.

\textit{Rule and Orbit-State (RO-S)}: given an orbit $\mathcal{O}^T(x)$ and the local rule $\rho$ predict the state $x^{(T+k)}$ at time $T+k$. Since the rule is explicitly provided, the model can bypass inference of rule structure and focus solely on learning to apply the update.

The rule in our 1dCA setup is based on a neighborhood radius $r = 2$, meaning each bit of the next state depends on a 5-bit window (2 left + current cell + 2 right) from the current state. Since there are $2^5$ possible 5-bit strings, the rule mapping can be represented by a 32-bit string. Each bit in this string corresponds to the output of the rule for a specific input. The position of this output bit within the rule string is determined by the binary value of the 5-bit input (see Fig.\ref{fig:methods}a). 
For evaluation, we use exact match for state prediction (1 if the full state is correct, 0 otherwise) and bit accuracy for rule prediction. We choose bit accuracy for the rule because exact rule matches are often uninformative: in many samples, not all $32 = 2^5$ transitions appear in the orbit, so the full rule cannot be uniquely recovered, biasing exact matches toward 0. See \autoref{appx:samples} for examples.

When tasked with predicting both future states and the underlying rules (O-RS setting), Figure~\ref{fig:combined_performance}(b) shows that models generally achieve high accuracy on rule prediction, though with interesting variations. ARMT notably struggles with accurate rule inference compared to other architectures, despite handling next-state prediction well.

Finally, we considered the setting in which the local rule $\rho$ is explicitly provided to the model, corresponding to the Rule and Orbit-State (RO-S) task. Intuitively, this should be the easiest version of the problem, since the model no longer needs to infer the rule from the orbit and can instead focus on applying the given rule to predict future states. As shown in ~\autoref{fig:combined_performance}(c), \gptneox indeed performs accordingly for next-state prediction, achieving near-perfect accuracy at $k=1$.

Surprisingly, however, this advantage does not persist for longer look-ahead horizons. For $k=2$, $3$, and $4$, performance drops to roughly the same level as in the original O-S setting, where the rule is not given explicitly. This suggests that providing the rule is sufficient for one-step prediction, but not for reliable multi-step state propagation.

\section{Samples examples}

\subsection{Handsup game}
\label{sec:handsup_sample}
\begin{lstlisting}[basicstyle=\ttfamily\scriptsize,breaklines=true, xleftmargin=0pt, frame=single]
You peek through a doorway into a cosy room.
7 friends sit around a round table in this order: Alice, Bob, Carol,
Dave, Erin, Frank, and Grace - and then back to Alice again.
They don't talk. At the end of each round they all decide, at the very
same moment, either to raise a hand or to keep both hands on the table.

You watch and jot down what happens:
- Round 1. Alice, Bob, Dave, Erin, Frank, and Grace raise their hands. The others keep their hands on the table.- Round 2. Alice, Carol, Erin, Frank, and Grace raise their hands. The others keep their hands on the table.- Round 3. Bob, Dave, Frank, and Grace raise their hands. The others keep their hands on the table.- Round 4. Alice, Carol, and Erin raise their hands. The others keep their hands on the table.- Round 5. Bob and Dave raise their hands. The others keep their hands on the table.
Now it's your turn to be the clever observer.
Puzzle: What will each friend do in Round 6?
Please answer in plain words, going in order around the table, starting from the first name above. Answer with the list of people with hands up, not mentioning the ones with hands down. For example: Alice, Bob, and Dave raise their hands.
\end{lstlisting}

\subsection{ECA - r2s20T10}
\label{appx:samples}

The input vocabulary of the tested models consists of the following tokens: \texttt{[0]}, \texttt{[1]}, and \texttt{[SEP]}. The states and the local rule $\rho$ are encoded as binary strings. The model receives the orbit as a sequence of bits, representing consecutive states separated by the \texttt{[SEP]} tokens.

We train the model to predict the blue tokens.

In all these examples rule is \texttt{01011111100100000101111011111100} and the initial state is \texttt{10110111001000110100}.

\textbf{O-S}

\noindent\texttt{10110111001000110100<sep>\\11101001101111101100<sep>\\10111011010000111011<sep> \\
11001110111011101100<sep>\\10111011001100111011<sep>\\11001110111011101100<sep>\\ 10111011001100111011<sep>\\11001110111011101100<sep>\\10111011001100111011<sep> \\ 11001110111011101100<gen>\\\color{blue}\textbf{10111011001100111011}
}

\textbf{O-O}

\noindent
\texttt{10110111001000110100<sep>\\11101001101111101100<sep>\\10111011010000111011<sep> \\
11001110111011101100<sep>\\10111011001100111011<sep>\\11001110111011101100<sep>\\ 10111011001100111011<sep>\\11001110111011101100<sep>\\10111011001100111011<sep> \\ 11001110111011101100<gen>\\\color{blue}\textbf{10111011001100111011<sep>\\11001110111011101100<sep>\\ 10111011001100111011<sep>\\11001110111011101100}
}

\textbf{O-RS}

\noindent
\texttt{10110111001000110100<sep>\\11101001101111101100<sep>\\10111011010000111011<sep> \\
11001110111011101100<sep>\\10111011001100111011<sep>\\11001110111011101100<sep>\\ 10111011001100111011<sep>\\11001110111011101100<sep>\\10111011001100111011<sep> \\ 11001110111011101100<gen>\\\color{blue}\textbf{01011111100100000101111011111100<sep>\\10111011001100111011}
}

\textbf{RO-S}

\noindent
\texttt{01011111100100000101111011111100<sep>\\10110111001000110100<sep>\\11101001101111101100<sep>\\10111011010000111011<sep> \\
11001110111011101100<sep>\\10111011001100111011<sep>\\11001110111011101100<sep>\\ 10111011001100111011<sep>\\11001110111011101100<sep>\\10111011001100111011<sep> \\ 11001110111011101100<gen>\\\color{blue}\textbf{10111011001100111011}
}

\section{Multiple Prediction Horizons Training}
\label{appx:adaptive_ca}
Given an orbit $\mathcal{O}^T(x)$ and the random shift token $s_i \in \{s_1, s_2,s_3,s_4\}$ the objective is to predict the state $x^{(T+i-1)}$. In this setup, we train the model to reason more for some inputs than others.

\begin{figure*}[t]
\centering
\includegraphics[width=0.7\textwidth]{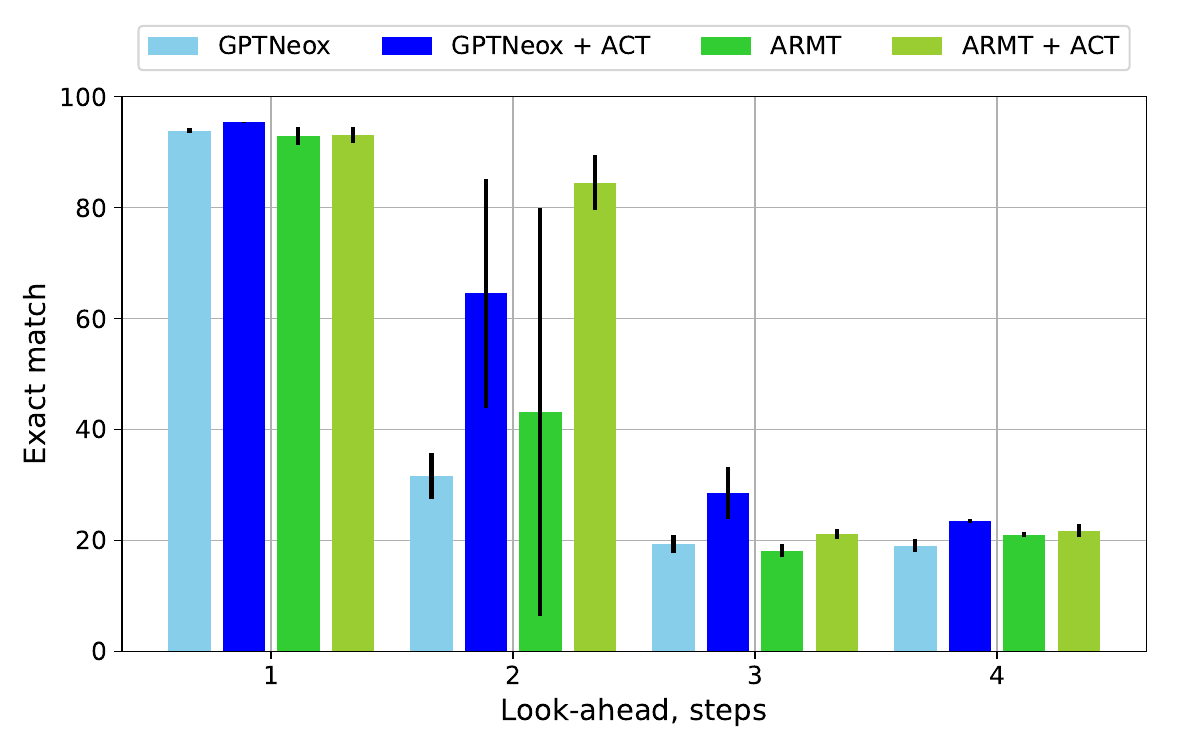}
\caption{\textbf{ACT outperforms the base model on multiple prediction horizons task}. Exact match accuracy (mean $\pm$ std) for cellular automata state prediction across different look-ahead horizons. Models receive initial 10 states followed by a special shift token (1-4) indicating prediction horizon. }
\label{fig:adaptive_ca }
\end{figure*}
We conducted experiments where a single model was trained to handle multiple prediction horizons (1-4 steps ahead) using special shift tokens in the input format: $\texttt{[x\_0][SEP]...[x\_9][shift\_k][gen][MASK]}$ where $k \in \{1,2,3,4\}$ indicates the required look-ahead. As shown in Figure \ref{fig:adaptive_ca }, baseline \gptneox performs 32\% shift=2 and 19\% for shift=4. Introducing ACT substantially mitigates these drops.
The ARMT architecture shows comparable characteristics -- while baseline performance at shift=2 is stronger than \gptneox (43\% vs 32\%), ACT provides similar absolute improvements (85\% at shift=2). However, both architectures exhibit similar limitations at the longest horizons (shift=4), with all variants scoring 21\%-25\%, indicating challenges in extreme-depth reasoning.

\section{Group Multiplication Task}
\label{appx:grmul}

Given a sequence of elements of some group, the task is to label each element with the product of all previous elements, including the current one \citep{merrill2024illusion, peng2025rwkv7gooseexpressivedynamic}. This task is relevant to reasoning as it provides a controlled setup with varying complexity. We evaluated our models on 3 groups of different difficulty: $Z_{60}$, $A_4 \times Z_5$, and $A_5$; and sequence lengths: 5, 10, 15, 20, and 40. For each model, we report the minimum number of layers required to achieve 70\% exact-match accuracy. 
For consistency with prior work, we use $d_{\text{model}} = 512$ and $n_{heads} = 8$. For ARMT, we use segments of size 2.

\begin{figure*}[htp]
    \centering
    \includegraphics[width=\textwidth]{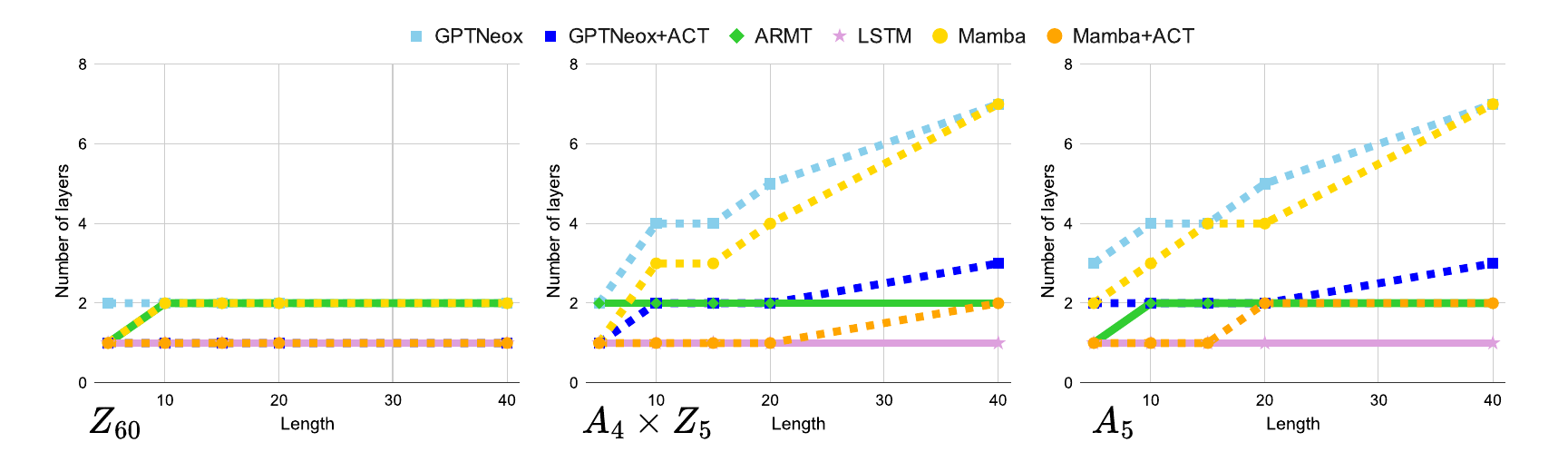}
    \caption{\textbf{ACT significantly reduces the required models' depth for the majority of group multiplication tasks.} Each chart contains the information about the minimal required number of layers for solving task of given length with 70\% exact match accuracy. GPTNeox and Mamba being $TC^0$-limited models require more layers for solving deeper (longer in this case) tasks, while ARMT and LSTM solve them with constant number of layers.}
    \label{fig:groupmul}
\end{figure*}

As shown in \autoref{fig:groupmul}, the required depth for solving longer tasks grows steadily for GPTNeox and Mamba models, while staying nearly constant (1--2 layers) for models with recurrence (ARMT and LSTM). We can further see that depth requirements can be significantly reduced by adding Adaptive Computation Time (ACT) or Associative Memory (ARMT), which is consistent with our findings on the 1dCA benchmark and highlights the benefits of effective depth scaling. LSTMs, however, perform much better, as they are able to solve the problem with just a single layer.

\section{Ablation Studies}
ACT was originally applied to single-layer NNs \citep{dehghani2018universal-transformer, graves2016adaptive}. For deep models, it can be applied either to each layer, averaging remainders across layers as a time penalty in the loss (layer-wise ACT, or LACT), or to the backbone model as a whole (MACT), mapping $\mathbb{R}^{N \times d} \rightarrow \mathbb{R}^{N \times d}$ without the embedding and unembedding layers. In our ablations, LACT and MACT perform similarly (see \ref{appx:mact}), so we use only layer-wise ACT in the main experiments.

To test whether gains come from adaptive computation or simply from more computation, we include a fixed-computation-time (FCT) baseline in our ablations (see \ref{appx:fct}). We use three fixed iterations, chosen to match the upper bound of the average number of ACT operations observed in our experiments.
Below, we present several auxiliary studies of ACT variants.


\subsection{Fixed Number of Steps in ACT vs Dynamic Number of Steps}
\label{appx:fct}

\begin{figure*}[h]
\centering
\includegraphics[width=0.8\textwidth]{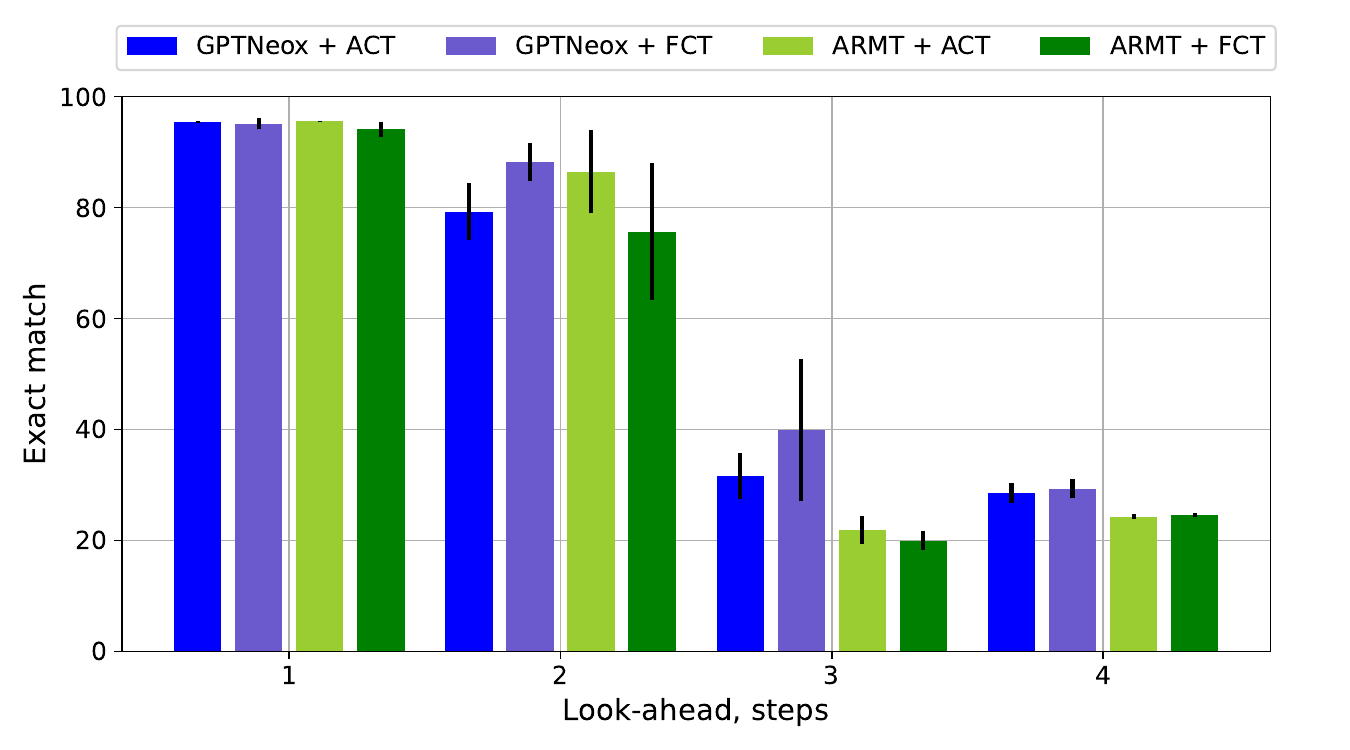}
\caption{\textbf{Fixed Computation Time (FCT) with 3 iteration steps performs on par with Adaptive Computation Time (ACT) in Orbit-State task.}
Exact match accuracy (mean ± std) for cellular automata state prediction across different look-ahead horizons.}
\label{ca-fct-o-s}
\end{figure*}

\begin{figure*}[h]
\centering
\includegraphics[width=0.8\textwidth]{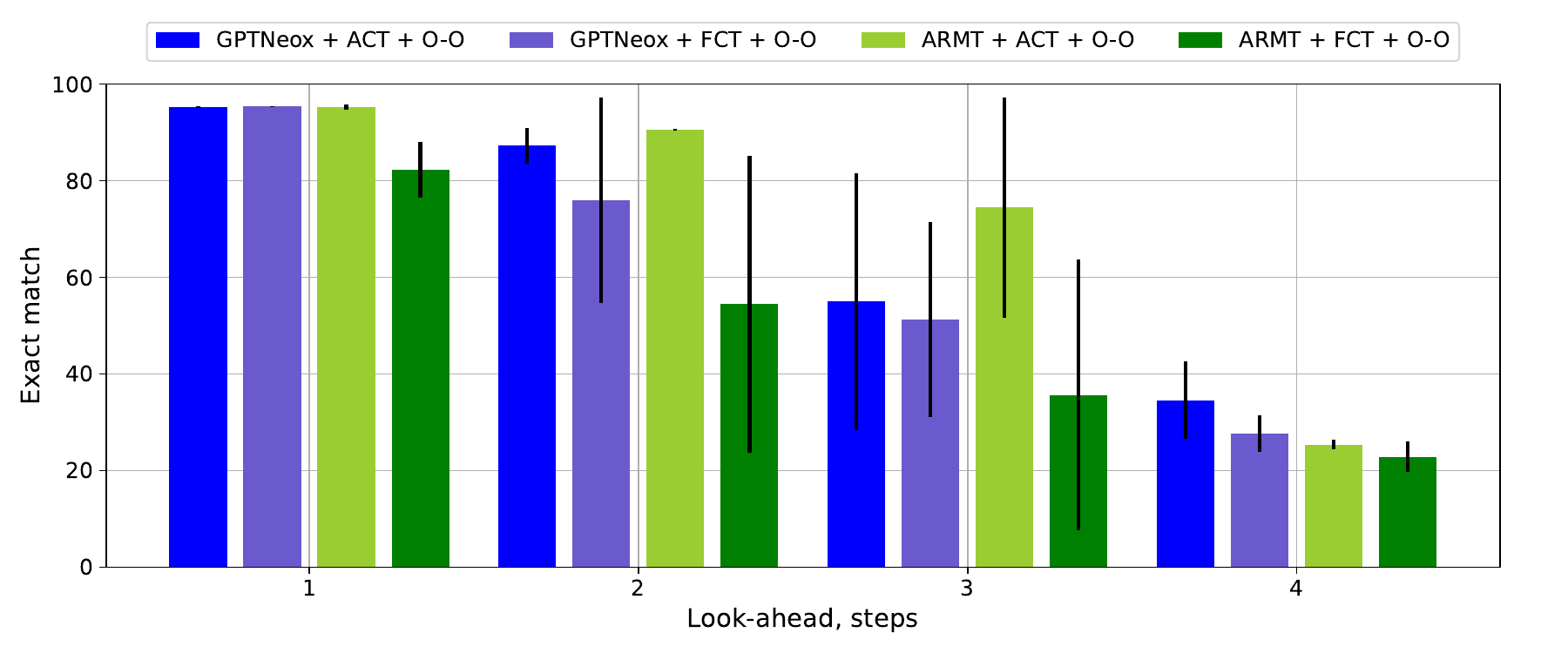}
\caption{\textbf{Fixed Computation Time (FCT) with 3 iteration steps underperforms Adaptive Computation Time (ACT) in Orbit-Orbit task.}
Exact match accuracy (mean ± std) for cellular automata state prediction across different look-ahead horizons.}
\label{ca-fct-o-o}
\end{figure*}

We conduct experiments with a fixed number of steps to assess the need for adaptivity in computation time.
A constant depth of 3 was selected based on experiments with ACT, which demonstrated that this represents the upper limit of the number of steps reached for any hidden state. The results with Fixed Computation Time (FCT) and ACT as the baseline are presented in \autoref{ca-fct-o-s} and \autoref{ca-fct-o-o} for O-S and O-O settings respectively.

In O-S setting, FCT improved the exact match in look-ahead 2, 3 for \gptneox, but performed worse in look-ahead 2 for ARMT.
In contrast, in the O-O setting, FCT showed reduced performance for both \gptneox and ARMT in look-ahead 2, 3, 4.
Therefore, adaptivity in computation time might find the optimal amount of steps leading to enhanced exact match, or perform equivalently with fewer steps.

\subsection{Model-ACT vs Layer-ACT}
\label{appx:mact}
ACT performs similarly to or better than Model-ACT. Model-ACT processes hidden states, as in the COCONUT model \citep{hao2024coconut}, by feeding them back into the input, so similar reasoning behavior is expected. A difference appears when these variants are applied to ARMT. However, training was stopped after 30,000 steps, and the MACT-augmented model may not have had enough time to fully converge. All models in this experiment followed these restrictions for a fair comparison.

\begin{figure*}[h]
\centering
\includegraphics[width=0.6\textwidth]{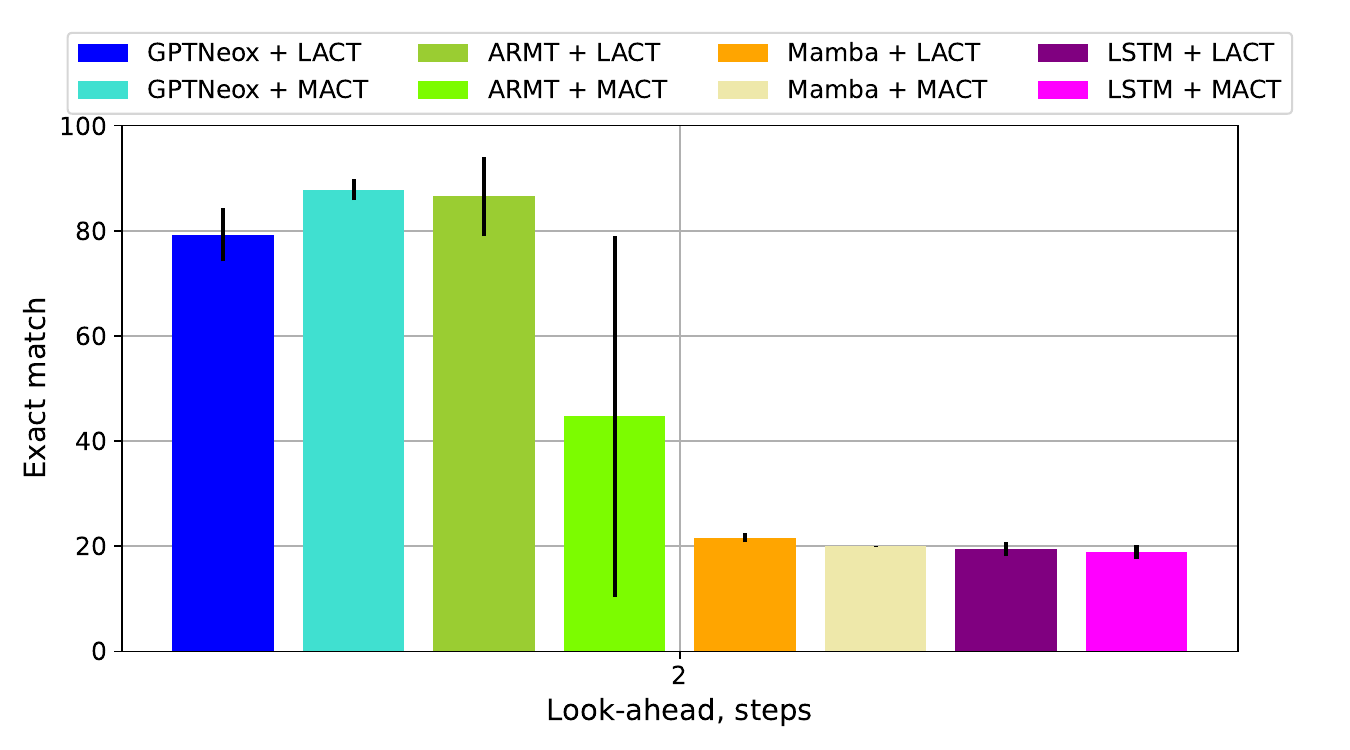}
\caption{\textbf{Layer-ACT performs similar or better compared to Model-ACT}. Exact match on cellular automata state prediction task with look ahead 2.}
\label{ca-act_cmp}
\end{figure*}

\end{document}